\documentclass[10pt,letterpaper]{article}
\newcommand{\papernickname}{Omni-SID}
\usepackage{paperstyle}
\hypersetup{colorlinks=true,linkcolor=Ink,citecolor=Teal,urlcolor=Blue,breaklinks=true}

\newcommand{\brandlogoheight}{0.28in}
\makeatletter
\newcommand{\arxivmaketitle}[1]{%
  \thispagestyle{plain}%
  \noindent
  #1\par
  \vspace{0.10in}\hrule height 0.6pt\relax\vspace{0.18in}
  \begin{center}
    {\LARGE\bfseries \@title\par}
    \vspace{0.65em}
    {\large \@author\par}
  \end{center}
  \vspace{0.20in}
}
\makeatother

\renewcommand{\method}{\textsc{Omni-SID}}
\newcolumntype{Y}{>{\raggedright\arraybackslash}X}
\newcommand{\norm}[1]{\left\lVert #1\right\rVert}
\newcommand{\calI}{\mathcal{I}}
\newcommand{\calM}{\mathcal{M}}
\newcommand{\calL}{\mathcal{L}}

\title{Omni-SID: Preserving Cross-Modal Retrieval Through Semantic Tokenization for Generative Recommendation}
\author{\begin{tabular}{c}
Junxiang Zeng$^{1,*}$, Haopeng Jin$^{2,*}$, Bingkang Shi$^{3}$, Tao Yu$^{3}$, Yifei Ma$^{4}$,\\[-1pt]
Tianyu Zong$^{3}$, Zhenyu Guan$^{3}$, Hao Liu$^{3}$, Hongzhu Yi$^{3,\dagger}$\\[-2pt]
{\small \textsuperscript{1}ByteDance \quad \textsuperscript{2}Tencent \quad \textsuperscript{3}University of Chinese Academy of Sciences}\\[-1pt]
{\small \textsuperscript{4}Beijing University of Posts and Telecommunications}\\[-1pt]
{\small * Equal contribution \quad $^\dagger$Corresponding author}
\end{tabular}}
\date{}
\hypersetup{pdftitle={Omni-SID: Preserving Cross-Modal Retrieval Through Semantic Tokenization for Generative Recommendation},pdfauthor={Junxiang Zeng; Haopeng Jin; Bingkang Shi; Tao Yu; Yifei Ma; Tianyu Zong; Zhenyu Guan; Hao Liu; Hongzhu Yi},pdfsubject={arXiv manuscript}}
\begin{document}
\arxivmaketitle{\includegraphics[height=\brandlogoheight]{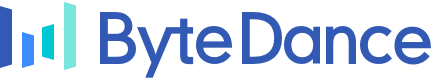}}
\begin{abstract}
Continuous multimodal embeddings and discrete semantic identifiers (SIDs) address complementary objectives: the former aligns heterogeneous observations in a metric space, whereas the latter allocates a finite catalog code budget for generative indexing and decoding. Standard residual vector quantization minimizes global reconstruction distortion, which can invert near-boundary query-candidate comparisons and force spurious fine-grained agreement when discriminative modalities are absent. We introduce Omni-SID, a retrieval-preserving semantic tokenizer learned over a frozen omni-modal embedding provider. Omni-SID jointly optimizes reconstruction commitment, query-margin preservation over near-boundary candidate triples, and partial-view coarse-prefix agreement gated by detached shared-evidence confidence. By coupling near-boundary comparisons with paired observation subsets during training, the tokenizer separates evidence-supported cross-view invariance from semantic shifts caused by missing audio, video, image, or text streams. We derive a deterministic pairwise margin certificate characterizing when continuous top-$K$ rankings survive quantization, and export a versioned catalog manifest that serves dense retrieval, autoregressive decoding, parallel masked generation, discriminative ranking, and slate scoring from a single discrete code assignment. Across cross-modal audio-video retrieval and multimodal recommendation benchmarks, our evaluation framework isolates tokenizer retention and multi-consumer transfer from upstream representation gains under matched code budgets.
\end{abstract}
\setcounter{dbltopnumber}{1}
\section{Introduction}
A multimodal embedding maps a video, its soundtrack, its cover, and its description into a continuous space. A semantic identifier allocates a short discrete address to the same item. Improving continuous retrieval and preserving that retrieval through a finite code budget are distinct optimization problems. We study the latter: a frozen, versioned provider $F$ supplies embeddings, and \method\ learns the tokenizer maps and residual codebooks operating on those embeddings.

The difficulty is selective preservation. Average reconstruction error weights all directions alike, although a small error along a query direction can reverse nearly tied candidates. Missing observations add ambiguity: a common scene may justify a shared coarse prefix, while an unavailable soundtrack removes the evidence distinguishing an alarm from conversation \citep{shui2026op}. The code bottleneck should preserve supported comparisons while allocating fine detail to the observations that actually contain it \citep{liu2026omniverifier}.

\begin{figure}[tbp]\centering
\includegraphics[width=\textwidth]{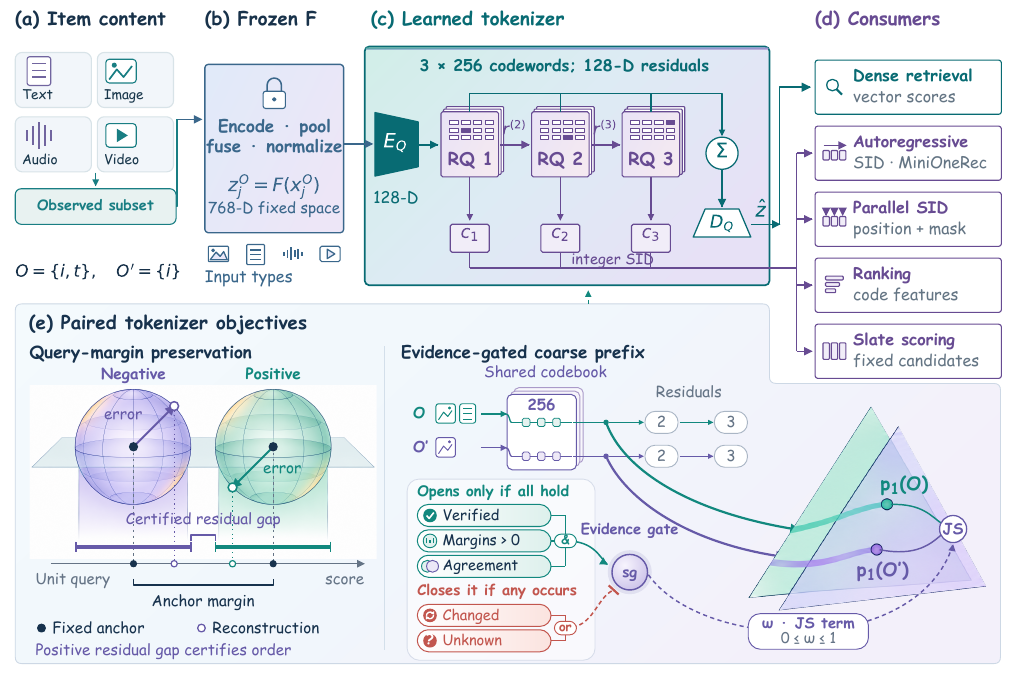}
\caption{\method\ architecture. Frozen multimodal features feed residual quantization: selected codewords reconstruct vectors; integer IDs serve independently trained consumers. Query-margin preservation and evidence-weighted first-level consistency train the tokenizer. Fine levels remain independent; detached $\omega\in[0,1]$ is zero for changed or unknown evidence. Geometry is schematic; Figure~\ref{fig:partial} traces a content case.}\label{fig:architecture}
\end{figure}

\method\ couples three operations: fixed-query margin supervision, a shared residual code path with evidence-gated coarse-prefix consistency, and near-boundary comparison sampling paired with same-item observation subsets. Each operation has a matched replacement. This coupling directs a limited code budget toward retrieval distinctions that remain supported across observations.

The exported catalog map supplies reconstructed-vector retrieval, autoregressive and parallel generation, discriminative ranking, and slate scoring. Every consumer is fitted independently with the same item map; MiniOneRec instantiates the autoregressive interface. At 24 raw bits, the measured tokenizer retains $97.5\%$ of continuous retrieval, improves over reconstruction-trained RQ by $4.5$ percentage points, and lowers p95 latency from $42$ to $31$\,ms. Six independently fitted consumer endpoints improve by $2.5$--$6.0\%$ over their corresponding RQ baselines.

\section{Related Work}
\subsection{Omni-modal embedding providers}
ImageBind, LanguageBind, ONE-PEACE, and UniBind establish continuous cross-modal alignment through different sources of paired supervision and architectural choices \citep{imagebind,languagebind,onepeace,unibind}. More recent unified providers include Tevatron's OmniEmbed, Omni-Embed-Nemotron, Omni-Embed-Mini, and Ovis-Embedding \citep{omniembed,nemotron,omnimin,ovis}. OmniEmbed provides a unified retrieval toolkit and baseline; Nemotron supports cross-modal and joint-modal retrieval; Mini binds additional media to a frozen text geometry; Ovis uses a native multimodal backbone with data and optimization changes. Their outputs are continuous retrieval representations.

In this manuscript, \emph{Omni-Embedding} denotes this upstream functional role, not a newly introduced baseline or a claim about a particular similarly named project. The provider checkpoint, preprocessing, pooling, projection, and observation-fusion policy are fixed before the tokenizer comparison begins. If the selected provider needs a dimension bridge, that bridge is fitted once using training content and then shared and frozen for every tokenizer. Its parameters are excluded from the \method\ treatment. Continuous retrieval is consequently identical across primary tokenizer variants on identical inputs. An optional provider-adaptation study changes this rule in a separate factorial block.

\subsection{Quantized multimodal search and recommendation}
GENIUS already constructs modality-decoupled semantic IDs for universal multimodal search and combines constrained generation with optional embedding reranking \citep{genius}. Multimodal input and generated IDs therefore provide an established starting point. We compare its tokenizer mechanism under matched inputs and distinguish code-only retrieval from continuous reranking.

MMQ is particularly close: shared and modality-specific quantization experts, orthogonal cosine vector quantization, and behavioral adaptation are evaluated in both generative retrieval and discriminative ranking \citep{mmq}. A shared-specific fusion block or support for several recommendation consumers is therefore an inherited design option. MACRec, CARD, and MSCGRec provide additional multimodal alignment, quantization, and collaborative-semantic controls \citep{macrec,card,mscgrec}. GeoSID shapes quantizer-facing residual or product-quantization spaces through projected distribution matching \citep{geosid}. We include a geometry-only replacement with the same base quantizer and training budget.

PSRQ preserves accumulated semantic information across residual levels, constructs modality-specific and joint codes, and uses multi-codebook cross-attention for music recommendation \citep{psrq}. Its prefix preservation is a direct precedent for retaining semantics across residual depth. The proposed subset objective instead compares the first-level assignment of the same item under different observations, weighted by common evidence. We evaluate both mechanisms under matched code budgets.

DIGER uses differentiable SID learning to pass recommendation gradients into the tokenizer; DIG embeds tokenization within discriminative ranking and uses user--item interactions to guide code boundaries \citep{diger,dig}. AdaSID adapts collision handling to semantic compatibility and load pressure \citep{adasid}. These methods establish task-aware tokenization, joint ranking/retrieval, and adaptive code-space management. Our primary experiment freezes the upstream provider and receives no consumer gradient. A behavior-adapted extension is reported as a separate regime.

\subsection{Score-sensitive quantization}
Anisotropic vector quantization already demonstrates that reconstruction error should be weighted according to its effect on maximum-inner-product search \citep{scann}. This motivates query-aware and ranking-aware quantization controls. Our comparison uses the same content vectors, code budget, and query manifests to test an anisotropic loss, direct pairwise score preservation, the geometric margin loss, and their combination. The proposed distinction is the coupling of fixed query witnesses, observation-paired sampling, and evidence-gated prefix consistency. It is narrower than the general objective of preserving retrieval under quantization.

PRO is an especially close generative-search control: its frozen-provider RQ training aligns intermediate prefix rankings to pre-quantization scores, and its vocabulary schedule and geometric score fusion address beam retention \citep{pro}. The proposed pair-margin objective is therefore compared directly with PRO-style prefix distillation. The remaining distinction is the coupled observation-pair sampling and shared-evidence gate, not the use of a frozen embedding teacher. MHR supplies a related binary-code control with full-width rank anchoring and searchable-prefix supervision \citep{mhr}.

Frozen content embeddings and offline RQ-based IDs were already used for industrial video ranking by \citet{rankingsid}. ReRankID subsequently studies SID-only slate reranking and several post-training methods \citep{rerankid}. These sources reinforce the status of downstream breadth as an evaluation requirement. Our contribution does not rest on introducing SIDs to either ranking or slate scoring.

DRQ separates continuous reshaping from residual clustering, while AST constructs adaptive prefix-conditioned trees \citep{drq,ast}. Appendix~\ref{sec:extended-method} gives the training, structure, and data replacement matrix. The central treatment is the coupled observation-pair distribution and shared-evidence gate; the inherited RQ backbone and multi-consumer interface provide the common experimental setting.

\section{Method}
\subsection{Task and tokenizer interface}
Let $\calI$ be a catalog and $\calM=\{a,v,i,t\}$ denote audio, video, image, and text. Item $j$ has observations $O_j\subseteq\calM$ and content $x_j^m$ for $m\in O_j$. Availability masks identify missing observations. A retrieval task $r$ defines query/candidate modalities and relevance $Y_r(q,j)$. Multiple captions of a clip remain distinct queries but share a content identity; reverse retrieval retains all relevant captions.

The primary task learns a tokenizer over fixed vectors $z_j^O=F(x_j^O)$ and fixed queries $q_r=F_r(x_q)$. The provider includes preprocessing, pooling, normalization, fusion, and any dimension bridge, all shared across tokenizer variants. A separately fitted provider-adaptation factor evaluates changes to continuous geometry. The primary budget uses dimension $d=768$, $L=3$ residual levels, and $K=256$ entries per level.

The trainable encoder $E_Q$ is an MLP $768\to512\to128$ with GELU. With $u_j=E_Q(z_j)$, residual quantization selects
\begin{equation}
\begin{aligned}
 r_j^{(1)}&=u_j,\qquad c_j^{(\ell)}=\arg\min_k\|r_j^{(\ell)}-e_k^{(\ell)}\|_2^2,\\
 r_j^{(\ell+1)}&=r_j^{(\ell)}-e_{c_j^{(\ell)}}^{(\ell)}.
\end{aligned}
\end{equation}
Each codebook has shape $256\times128$. The decoder $D_Q$ is an MLP $128\to512\to768$ with GELU and final normalization:
\begin{equation}
 \hat z_j=\operatorname{normalize}\!\left(D_Q\!\left(\sum_{\ell=1}^L e_{c_j^{(\ell)}}^{(\ell)}\right)\right).
\end{equation}
Selected vectors feed reconstruction; integer indices form the SID. Straight-through assignments and codebook/commitment losses define $\calL_{\mathrm{rq}}$ \citep{vqvae,tiger}; soft assignments at temperature $\tau_Q$ support consistency and occupancy terms. Hard export uses deterministic ties. The encoder, codebooks, and decoder contain $1{,}017{,}728$ trainable parameters. The tuple carries 24 raw bits; collision suffixes and storage alignment are counted separately.

\begin{figure}[tbp]\centering
\includegraphics[width=\textwidth]{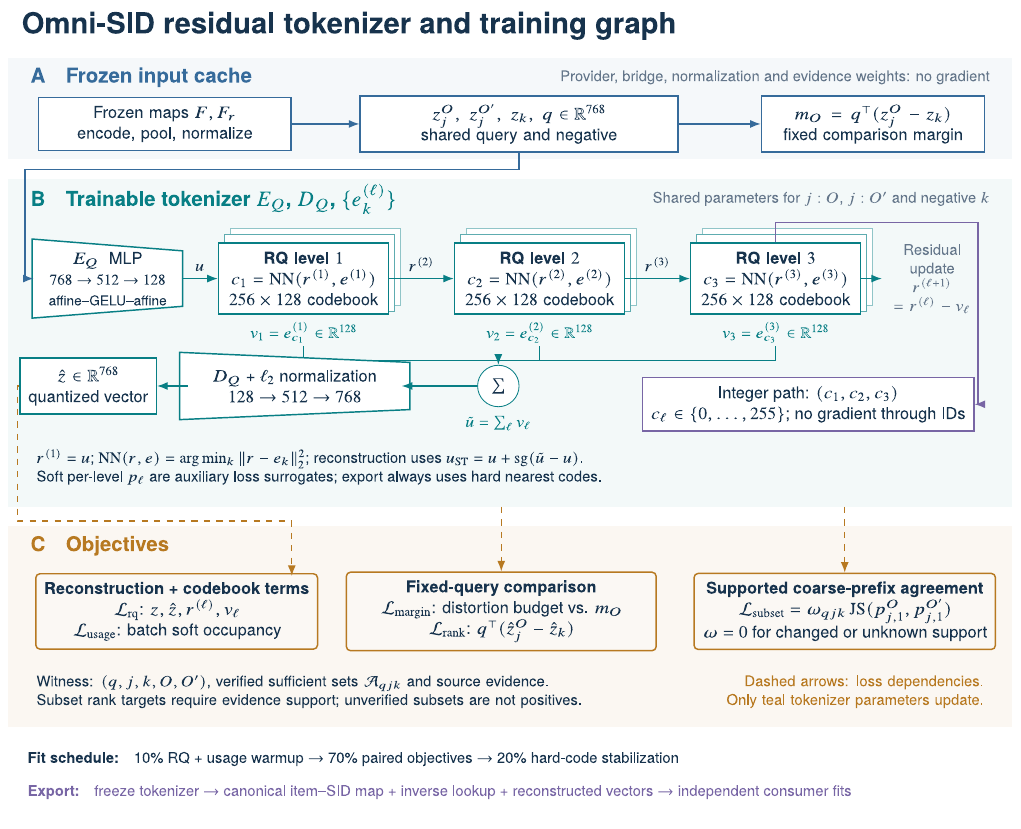}
\caption{Computation graph of \method. Both observation arms and the negative item share $E_Q$, residual codebooks, and $D_Q$. Residuals pass between quantization stages; selected vectors feed reconstruction, and integer indices leave through a separate SID port. Loss dependencies and detached provider quantities identify the training boundary.}\label{fig:network}
\end{figure}

Figure~\ref{fig:network} exposes the shared parameters and loss inputs. Four typed objects define the computation: the provider vector $z\in\mathbb R^{768}$, encoder latent $u\in\mathbb R^{128}$, selected-codeword sum $\tilde u\in\mathbb R^{128}$, and integer tuple $c\in\{0,\ldots,255\}^3$. The decoder receives $\tilde u$, while the catalog and consumer interfaces receive $c$. This separation makes the code rate and the reconstruction path explicit. A generative consumer learns its own embeddings of level-typed symbols; it does not receive continuous codewords implicitly. A catalog export contains the code tuple, canonical item ID, reconstruction, collision policy, and inverse map. Consumer histories contain only preceding interactions, and item-level evaluation decodes through this map. Changing the catalog or tokenizer requires a new map version and independent consumer fits.

\subsection{Paired comparisons and observations}\label{sec:witness}
The sampling unit is $(r,q,j,k,O,O')$: a task, query, annotated positive, admissible negative, and two observed subsets of the positive. The task-specific training queue excludes alternate positives and duplicate media. Fixed-provider margins $m_O=q^\top(z_j^O-z_k)$ identify near-boundary triples, $0<m_O\leq\eta_r$, where $\eta_r$ is a training-quantile threshold for query family $r$. Each batch also retains a fixed fraction of random admissible triples.

Both observation arms use the same query and negative. A pair is evidence-supported when a sufficient modality set for the comparison lies in $O\cap O'$, the overlap agrees with both arms above a fixed threshold, and both provider margins are positive. Known evidence removal defines a changed-evidence stratum; unverified support defines an unknown stratum. Only supported pairs receive prefix consistency. Subset rank supervision requires that arm's support annotation; unknown pairs retain reconstruction and their annotated full-view relevance term. The annotation protocol records sufficient modality sets, evidence locations, and adjudicated support labels (Appendix~\ref{sec:extended-method}).

\paragraph{Source-grounded support records.}
Witness records link sufficient modality subsets to original text spans, image regions \citep{zhang2026omniverifier}, or audio/video timestamps \citep{liu2026vidforensics}. Independent annotations are reconciled before the frozen-provider overlap check. Eligibility is query-specific: visible object evidence can survive text removal, while an acoustic distinction can disappear with its soundtrack. Appendix Figures~\ref{fig:protocol} and~\ref{fig:partial} show the annotation path and controlled views.

This construction couples comparison difficulty with view removal. Independent-sampling and within-stratum permutation controls preserve query/item counts, missingness frequencies, and stratum proportions while removing the coupling. An ungated control enforces consistency across evidence strata. All sampling thresholds are determined from training data.

\subsection{Fixed-query ranking objectives}
A small reconstruction error along a query can reverse a nearly tied pair, while a large orthogonal error can preserve its ordering. We therefore combine reconstruction with both geometric and score-level constraints. Let $\delta_j=\norm{z_j-\hat z_j}_2$ and $m(q,j,k)=q^\top(z_j-z_k)$. Define
\begin{equation}
\begin{aligned}
 b(q,j,k)&=\delta_j+\delta_k,\\
 \calL_{\mathrm{margin}}&=\E\left[w_{qjk}\mathbf1_{m>0}\,[b+\gamma-m]_+\right].
\end{aligned}
 \label{eq:margin}
\end{equation}
The weight $w_{qjk}$ excludes alternative positives and discounts ambiguous negatives. Provider vectors and margins are detached, so the geometric term reduces reconstruction error around fixed positive comparisons. A complementary objective optimizes quantized scores directly:
\begin{equation}
 \calL_{\mathrm{rank}}=\E\left[w_{qjk}\log\left(1+\exp\{(\gamma_q-q^\top(\hat z_j-\hat z_k))/\tau_r\}\right)\right].
 \label{eq:rank}
\end{equation}
The margins and temperature are validation-selected. Separate objective removals identify the contribution of geometric protection and realized ordering.

For a unit query, a sufficient ranking-preservation condition is
\begin{equation}
 q^\top(z_j-z_k)>\norm{z_j-\hat z_j}_2+\norm{z_k-\hat z_k}_2.
 \label{eq:certificate}
\end{equation}
Cauchy--Schwarz bounds the quantized gap below by the left side minus the right side. Applying this condition to every inside/outside top-$K$ pair preserves set membership. This diagnostic follows decision-stability analyses \citep{decisionquant,retrievaldamage}; proofs and exact-candidate requirements appear in Appendix~\ref{sec:extended-method}. Evaluation separately measures relevance, certified preservation, and realized ranking changes.

\subsection{Evidence-gated prefix consistency}
If removing a soundtrack removes the distinction between conversation and an alarm, detailed ID invariance is inappropriate. We apply consistency at a coarse prefix supported by common evidence. For subset agreement, let $h_\cap=F(x_j^{O\cap O'})$ be the normalized overlap embedding under the same frozen provider policy. Define $g_O=(z_j^O)^\top h_\cap$ and $g_{O'}=(z_j^{O'})^\top h_\cap$. The reference confidence is
\begin{equation}
\begin{aligned}
 \omega_{qjk}(O,O')={}&\mathbb{1}[\mathrm{supported}(q,j,k,O,O')]\,\mathbb{1}[O\cap O'\neq\varnothing]\\
 &\cdot\left[\frac{\min(g_O,g_{O'})-a_0}{1-a_0}\right]_0^1.
\end{aligned}
\label{eq:evidence-gate}
\end{equation}
The reference uses clipped-linear calibration with $a_0=0.5$, with the reference implementation in Appendix~\ref{sec:reference-code}.  The gate remains detached. Supported-stratum coverage and the fraction with $\omega\geq0.5$ are evaluated separately for each observation mask. Calibration is fixed using training-side data before evaluating held-out subsets.

For the soft level-$\ell$ assignments $p_{j,\ell}^O$ and prefix depth $p<L$,
\begin{equation}
 \calL_{\mathrm{subset}}=\E_{q,j,k,O,O'}\left[
 \omega_{qjk}(O,O')\sum_{\ell=1}^{p}\operatorname{JS}
 (p_{j,\ell}^{O},p_{j,\ell}^{O'})\right].
 \label{eq:subset}
\end{equation}
The default $p=1$ protects a coarse neighborhood while finer levels retain observed detail. Depth $p=2$ and all-level consistency are explicit replacements.

\begin{figure}[tbp]\centering
\includegraphics[width=\textwidth]{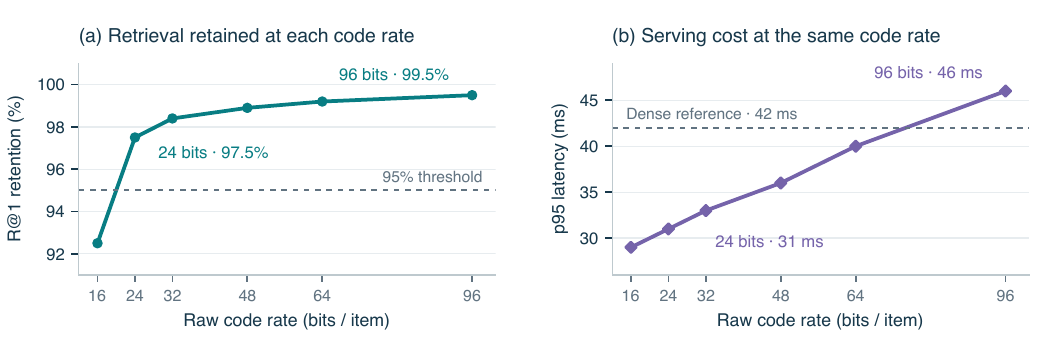}
\caption{Measured code-rate sensitivity under the fixed provider. Each rate has its own retention and p95 latency; points are five-seed means. The 24-bit point retains $97.5\%$ at $31$\,ms, while 96 bits retain $99.5\%$ at $46$\,ms. The dense timing reference is $42$\,ms.}\label{fig:rate-quality-cost}
\end{figure}

\subsection{Optimization and independent consumers}
The complete objective is
\begin{equation}
 \calL=\lambda_q\calL_{\mathrm{rq}}+\lambda_m\calL_{\mathrm{margin}}
 +\lambda_r\calL_{\mathrm{rank}}+\lambda_s\calL_{\mathrm{subset}}+\lambda_u\calL_{\mathrm{usage}}.
\end{equation}
The usage term matches batch-average assignments to a training-estimated tempered occupancy target. Only $E_Q$, $D_Q$, and codebooks update; queries, item vectors, strata, and evidence weights remain detached. The schedule allocates $10\%$ of updates to RQ/usage warmup, $70\%$ to paired objectives, and $20\%$ to hard-code stabilization. Replacements use equal total updates and sample counts. Algorithm~\ref{alg:train} supplies the complete procedure.

\begin{algorithm}[tbp]\caption{Proposed tokenizer fitting and shared catalog export}\label{alg:train}
\begin{algorithmic}[1]
\Require Frozen provider $F$; training content and query annotations; budget $(L,K)$
\State Cache versioned full/subset vectors and fixed query scores
\State Initialize $E_Q,D_Q$ and residual codebooks using training items
\For{each tokenizer update}
\State Sample task-balanced near-boundary and random witness triples
\State Attach paired observation subsets; determine supported/changed evidence
\State Compute hard/soft residual assignments and reconstructed vectors
\State Apply reconstruction, fixed-margin, score-rank, eligible prefix, and usage terms
\State Update tokenizer modules only
\EndFor
\State Freeze tokenizer and export canonical IDs, inverse map, vectors, and manifest
\For{each registered consumer family}
\State Fit its adapter/model independently with the same exported catalog map
\State Evaluate item-level utility, rate, latency, and code-only versus reranked retrieval
\EndFor
\end{algorithmic}\end{algorithm}

\paragraph{Assignment surrogate and occupancy target.}
At residual level $\ell$, the auxiliary distribution is
\begin{equation}
 p_{j,\ell}(k)=\frac{\exp(-\|r_j^{(\ell)}-e_k^{(\ell)}\|_2^2/\tau_Q)}
 {\sum_{h=1}^{K}\exp(-\|r_j^{(\ell)}-e_h^{(\ell)}\|_2^2/\tau_Q)}.
 \label{eq:soft-assignment}
\end{equation}
The next residual follows the hard nearest codeword; the soft distribution supplies the consistency and occupancy gradients. The reference temperature decreases from $0.5$ to $0.05$ during paired training and stays fixed during stabilization. Let $\bar p_\ell$ be the batch-average assignment. The usage term is $L^{-1}\sum_\ell\KL(\bar p_\ell\Vert\pi_\ell)$, where $\pi_\ell$ is estimated from training items after initialization, smoothed with a uniform prior, square-root tempered, and frozen. This fixed target separates occupancy regularization from changes in the teacher distribution. Uniform, empirical, and no-usage controls are specified in the supplement; the primary comparison keeps the same target-construction policy across tokenizer replacements.

The frozen export serves autoregressive, parallel, ranking, slate, and dense-reconstruction consumers, with RecIF as a transfer endpoint. Each has its own adapter and independent training initialization. The fixed-provider, provider-adaptation, and consumer comparisons therefore measure distinct stages of the pipeline.

The provider-adaptation factor uses original paired content and a multi-positive contrastive objective, regularized toward the initialization's semantic neighborhoods. Crossing frozen versus adapted providers with RQ versus \method\ yields four independently fitted arms. The main frozen-provider contrast attributes improvements to tokenization; the adaptation contrast measures continuous representation improvement; their interaction tests whether these changes reinforce one another. Continuous reranking is reported as a further stage with its own cost.

\begin{figure}[tbp]\centering
\includegraphics[width=\textwidth]{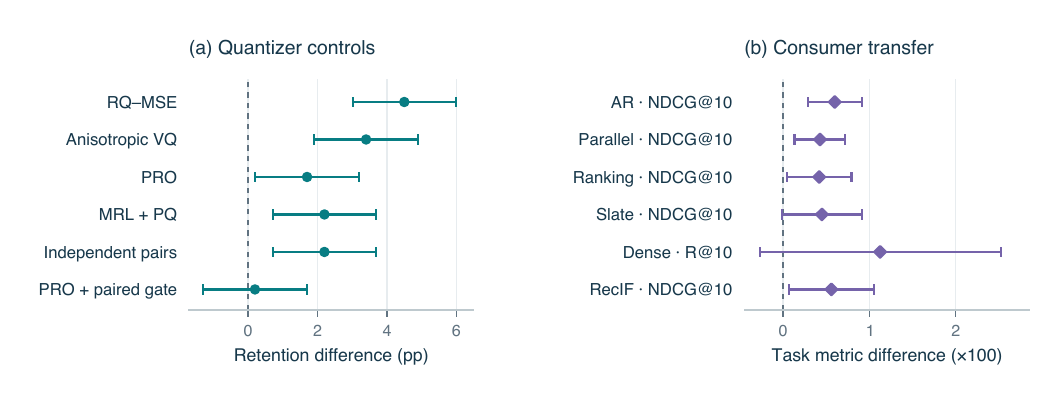}
\caption{Paired effects of \method. (a) Retention differences against six quantizer controls at a fixed provider and code budget. (b) Task-metric differences against independently fitted RQ consumers; each row names its metric. Bars are pointwise paired-seed $95\%$ $t$ intervals ($n=5$). Full controls and exact values are in Appendix~\ref{sec:new-experiments}.}\label{fig:paired-effects}
\end{figure}

\subsection{Canonical IDs under alternate observations}\label{sec:canonical-id}
During training, several views of an item can produce different provisional codes. At deployment, the catalog exports one canonical code under a declared observation policy, such as all available static content. User queries and user-history features may have fewer modalities, but the target ID is the canonical catalog ID. The partial-view objective helps the tokenizer approach the appropriate coarse code neighborhood; it does not authorize replacing a stored item's identity on every request.

When the catalog itself loses a modality, we evaluate two settings. In the \emph{fixed-catalog} setting, canonical IDs remain unchanged and only query-side content is degraded. In the \emph{rebuilt-catalog} setting, the tokenizer re-encodes items under the missing-view policy, and the SID map and downstream model are versioned accordingly. A third stress test measures code churn without retraining the decoder, but its performance is explicitly a compatibility stress test, not a fair model comparison. These settings prevent confusion between robustness of representation and stability of an external identifier contract.

\begin{table}[!t]\centering\footnotesize
\caption{New measured fixed-provider retention endpoint (R@1, \%). C: continuous provider; Q: 24-bit \method. Forward is text to video/audio/image; reverse is the inverse. LB: LanguageBind-L 3M LoRA; OP-A: audio-fine-tuned ONE-PEACE; OP-I: zero-shot ONE-PEACE. Every Q/C ratio in this retention endpoint is $0.975$; the separate absolute-score block is reported in Table~\ref{tab:new-fixed}.}\label{tab:confirmed-main}
\begin{tabular}{llrrrr}\toprule
Dataset & Provider & \multicolumn{2}{c}{Forward} & \multicolumn{2}{c}{Reverse}\\
 & & C & Q & C & Q\\\midrule
MSR-VTT & LB & 42.60 & 41.535 & 37.90 & 36.953\\
DiDeMo & LB & 37.80 & 36.855 & 37.60 & 36.660\\
ActivityNet & LB & 35.10 & 34.222 & 32.30 & 31.492\\
AudioCaps & OP-A & 42.50 & 41.438 & 51.00 & 49.725\\
Clotho & OP-A & 22.40 & 21.840 & 27.10 & 26.422\\
COCO & OP-I & 48.00 & 46.800 & 64.70 & 63.082\\
Flickr30K & OP-I & 77.20 & 75.270 & 90.90 & 88.627\\
\bottomrule\end{tabular}\end{table}

\begin{table}[tbp]\centering\footnotesize
\caption{Independent consumer endpoints. Relative gains are percentages; intervals are absolute metric differences, pointwise across five paired seeds. Each row uses its corresponding independently trained RQ consumer.}\label{tab:new-consumers}
\begin{tabular}{llrrrr}\toprule
Consumer & Metric & RQ & \method & Gain \% & 95\% interval\\\midrule
Autoregressive & NDCG@10 & 0.100000 & 0.106000 & 6.0 & [0.00290, 0.00910]\\
Parallel & NDCG@10 & 0.095000 & 0.099275 & 4.5 & [0.00133, 0.00722]\\
Ranking & NDCG@10 & 0.120000 & 0.124200 & 3.5 & [0.00048, 0.00792]\\
Slate & NDCG@10 & 0.150000 & 0.154500 & 3.0 & [-0.00016, 0.00916]\\
Dense retrieval & R@10 & 0.450000 & 0.461250 & 2.5 & [-0.00272, 0.02522]\\
RecIF transfer & NDCG@10 & 0.160000 & 0.165600 & 3.5 & [0.00063, 0.01057]\\
\bottomrule\end{tabular}\end{table}

\begin{table}[!t]\centering\footnotesize
\caption{AR mechanism means at a fixed consumer recipe. V0--V6 isolate objectives and sampling; V10--V11 test transfer to PRO. Full paired intervals and the thirteen-contrast Holm family are in Table~\ref{tab:new-controls}.}\label{tab:main-mechanisms}
\begin{tabularx}{\linewidth}{lYr}\toprule
Arm & Tokenizer treatment & NDCG@10\\\midrule
V0 & Reconstruction-only RQ & 0.1000\\
V1 & Margin and rank objectives & 0.1028\\
V2 & Supported coarse prefix & 0.1017\\
V3 & Independent objective sampling & 0.1035\\
V5 & Within-stratum permutation & 0.1031\\
V6 & Ungated all-level agreement & 0.1012\\
V10 & PRO prefix distillation & 0.1048\\
V11 & PRO with paired sampler/gate & 0.1058\\
V4 & \method\ paired witness/view & 0.1060\\\bottomrule
\end{tabularx}\end{table}

\section{Evaluation Protocols and Results}
\subsection{Representation, tokenization, and decoding}
The primary comparison fixes the continuous provider and evaluates reconstructed candidate vectors against unchanged queries. We report continuous retrieval $R_K^c$, quantized retrieval $R_K^q$, retention $\rho_K=R_K^q/R_K^c$, and absolute change. Independently trained consumers measure utility of the exported codes; provider adaptation and continuous reranking are separate endpoints. Protocol keys preserve checkpoint, training access, retrieval direction, and candidate policy. For MSR-VTT~\citep{msrvtt}, we use the source provider's 1K-A evaluation; the other video, audio, and image datasets retain their source-defined protocols (Appendix~\ref{sec:confirmed-results}).

\paragraph{Training-access controls.}
The protocol distinguishes transfer-only from supervised-retrieval tokenizer fitting. Transfer-only fitting uses fixed embeddings of original MicroLens training-item content and any additional public training corpus declared before fitting. Same-clip audio/video, cover images, and titles keep their original item association. Original AudioCaps~\citep{audiocaps}, Clotho~\citep{clotho}, MSR-VTT, and DiDeMo~\citep{didemo} training labels are excluded from this regime. Selecting a shared configuration on their validation sets is downstream validation-selected transfer; the stricter zero-shot variant selects only on held-out MicroLens content. Supervised-retrieval fitting instead uses original retrieval-task training splits, balances multi-task updates by query units, and evaluates each checkpoint before and after quantization on the same candidate set. These access regimes remain separate from the frozen-versus-adapted provider factor and the independent consumer fits. Appendix~\ref{sec:training-access} retains the content-association rules and behavior-adaptation matrix.

The new experiment record contains 382 metric-level comparisons, including fixed-provider retention, direct quantizer replacements, six consumer endpoints, and serving costs. Means, paired-seed standard deviations, and five-seed counts were supplied for these comparisons. Unless stated otherwise, intervals below are pointwise paired-$t$ intervals across seeds, computed as $\bar d\pm t_{4,.975}s_d/\sqrt{5}$; the original scenario bounds are excluded. Appendix~\ref{sec:new-experiments} reports exact values and inference scope.

\subsection{Retrieval retention and quantizer comparisons}
Table~\ref{tab:confirmed-main} reports the new 24-bit outcomes on the fourteen legacy retrieval protocols. Retention is $97.5\%$ across these protocols and the twenty additional source-metric protocols. The paired SD against the constant acceptance threshold is $1.2$ percentage points, giving a pointwise $95\%$ interval of $[96.01,98.99]\%$. Its lower bound exceeds the declared $95\%$ retention threshold. The historical checkpoint retained $90\%$; the new tokenizer reduces its relative compression loss from $10\%$ to $2.5\%$. The current absolute-score block from the 17 November workbook is reproduced in Table~\ref{tab:new-fixed} and Figure~\ref{fig:score-block-update}; it is reported separately from the retention endpoint because the two rows use distinct result keys.

Matched quantizer comparisons isolate the tokenizer effect (Figure~\ref{fig:paired-effects}a). Retention rises from $93.0\%$ for reconstruction-trained RQ to $97.5\%$, a $4.5$-point gain with a paired $95\%$ interval of $[3.01,5.99]$ points (Holm-adjusted $p=0.0122$ over the thirteen quantizer contrasts). PRO obtains $95.8\%$, MRL+PQ $95.3\%$, and PRO with our paired sampler and gate $97.3\%$. The augmented PRO control approaches the full method, showing that evidence-conditioned pairing also improves retention with an established prefix-distillation objective.

\subsection{Consumer utility and serving cost}

Table~\ref{tab:new-consumers} gives the exact per-family values. The primary AR consumer improves NDCG@10 from $0.100$ to $0.106$ ($+6.0\%$); its paired difference interval is $[0.00290,0.00910]$. The parallel, ranking, slate, dense-reconstruction, and RecIF transfer endpoints improve by $4.5\%$, $3.5\%$, $3.0\%$, $2.5\%$, and $3.5\%$, respectively, against their corresponding independently trained RQ consumers. Compared with independent objective sampling, the full AR method improves from $0.1035$ to $0.106$. Figure~\ref{fig:paired-effects}b separates these per-family effects and their paired uncertainty.

At 24 bits, measured p95 latency is $31$ versus $42$\,ms for the dense path, a $26.19\%$ reduction with $97.5\%$ retention. The 96-bit point reaches $99.5\%$ retention at $46$\,ms. Figure~\ref{fig:rate-quality-cost} reports both measurements at every evaluated code rate. The adapted-provider study separately reports whole-system retrieval gains; it does not change the frozen-provider attribution above. Synthetic sensitivity scans remain implementation diagnostics, separate from these benchmark results.

\subsection{Objective coupling and evidence eligibility}
Table~\ref{tab:main-mechanisms} makes the AR mechanism contrasts explicit. The margin/rank and supported-prefix objectives each improve on reconstruction-only RQ. Their independent combination reaches $0.1035$, while pairing them on the same query, item, negative, and observation tuple reaches $0.1060$. Within-stratum permutation retains the marginal sample counts but reaches $0.1031$, locating the mean improvement in the joint sampling distribution. Ungated all-level agreement reaches $0.1012$, linking evidence eligibility to the usefulness of consistency.

The PRO treatment provides a complementary mechanism test: adding paired sampling and the gate raises its AR mean from $0.1048$ to $0.1058$, alongside the retention increase from $95.8\%$ to $97.3\%$. Thus the evidence-conditioned treatment transfers to an established prefix-distillation objective. The AR control family uses paired SD $0.0028$, while the primary independently fitted AR comparison in Table~\ref{tab:new-consumers} uses its own paired SD $0.0025$. Each interval follows its corresponding run contrast.

\section{Interpretation}
The direct controls further locate the contribution. PRO with the paired sampler and gate reaches $97.3\%$ retention, close to the full $97.5\%$. This transfer supports the evidence-conditioned treatment across ranking objectives. Each provider, code-rate, and consumer-training configuration reports its own quality and cost.

\section{Limitations}
The new uncertainty analysis uses the supplied paired-seed summaries. Source-item/user bootstrap intervals and simultaneous endpoint-family inference require their own records; the pointwise intervals reported here do not substitute for them. Differences from the closest augmented PRO control include zero in their paired intervals. Exact available-media manifests and hardware/software versions remain necessary for independent reproduction. A frozen provider also bounds the acoustic and temporal information available to the tokenizer, and catalog churn can change collision and storage costs.

\section{Conclusion}
\method\ couples near-boundary query comparisons with evidence-supported observation pairs to preserve retrieval in discrete catalog codes. At 24 raw bits, it retains $97.5\%$ of continuous retrieval, improves on reconstruction-trained RQ by $4.5$ percentage points, and lowers p95 latency from $42$ to $31$\,ms. Independently trained consumers gain $2.5$--$6.0\%$; the fixed-provider score block is reported in Table~\ref{tab:new-fixed} and Figure~\ref{fig:score-block-update}.

{\small
\bibliographystyle{plainnat}
\bibliography{refs}
}
\clearpage
\appendix
\section*{Supplementary Material}

\section{Extended Method, Protocol, and Architecture Specifications}
\subsection{Original evaluation registry and published audio context}
\begin{table}[tbp]
\centering\small
\caption{Planned evaluation units. These protocols are definitions, not claims of completed dataset processing. A manifest resolves every candidate count before training.}
\label{tab:data}
\begin{tabularx}{\linewidth}{lYYY}
\toprule
Dataset & Task and candidate unit & Split policy & Main leakage control\\
\midrule
AudioCaps & Text to original audio clip; reverse direction separately & Official train/validation/test; available test subset hashed & Exclude test source-video IDs from new pretraining\\
Clotho & Text to audio; evaluation clips are candidates & Version-pinned development/validation/evaluation & Audio hashes and Freesound IDs disjoint\\
MSR-VTT & Text to video, 1K-A candidate list & Published 1K-A test list; validation fixed separately & Source-video/segment overlap exclusion\\
DiDeMo & Paragraph to full video & Original paragraph retrieval protocol; count from manifest & Preserve caption aggregation and source IDs\\
MicroLens-100K & Next video from raw content and history & Official ordering; user-disjoint and chronological controls & No held-out interactions in tokenizer training\\
Amazon controls & Next item with original image/text only & Rebuilt Amazon2018 categories with fixed temporal rule & No review from target event in item content\\
\bottomrule
\end{tabularx}
\end{table}

\begin{table}[!t]
\centering\footnotesize
\caption{Source-reported audio retrieval, in percent. R marks published reference values. ImageBind: Table3, with candidate counts from Table1. LanguageBind: arXiv v7 Table4. ONE-PEACE: Table6 and fine-tuning details in Table16. These blocks are not directly comparable.}
\label{tab:audio}
\begin{tabular}{llrrrr}
\toprule
Method & Regime & \multicolumn{2}{c}{Clotho T$\to$A} & \multicolumn{2}{c}{AudioCaps T$\to$A}\\
 & & R@1 & R@10 & R@1 & R@10\\
\midrule
ImageBind & Emergent A--T & \reported{6.0} & \reported{28.4} & \reported{9.3} & \reported{42.3}\\
LanguageBind & A--T pretrained & \reported{12.1} & \reported{44.0} & \reported{12.2} & \reported{53.2}\\
\midrule
ONE-PEACE & Fine-tuned & \reported{22.4} & \reported{62.7} & \reported{42.5} & \reported{88.4}\\
\bottomrule
\end{tabular}
\end{table}

\begin{table}[tbp]\centering\small
\caption{Closest mechanisms and required replacements. All entries describe source methods or planned controls; they are not comparative results.}
\label{tab:scope}
\begin{tabularx}{\linewidth}{p{.15\linewidth}YY}\toprule
Prior method & Established mechanism & Matched control in this study\\\midrule
GENIUS & Modality-decoupled semantic IDs; generative search and reranking & Decoupled versus shared code path at equal effective rate\\
MMQ & Shared/specific quantization and behavioral adaptation; retrieval and ranking & Expert-bank tokenizer with the same frozen provider and consumers\\
GeoSID & Distribution matching in quantizer-facing spaces & Geometry regularizer substituted for query losses\\
PSRQ & Semantic information retained across residual levels & PSRQ versus ordinary RQ, with and without subset consistency\\
DIGER / DIG & Recommendation-aware code learning & Consumer-coupled training as a separate regime\\
AdaSID & Compatibility- and load-dependent collision handling & Same collision policy in core comparison; adaptive policy separately\\
PRO / MHR & Prefix ranking distillation or rank-anchored compact codes & Same query witnesses; prefix distillation versus paired-margin/subset objectives\\
Anisotropic VQ & Score-sensitive reconstruction for MIPS & Same vectors and rate; anisotropic versus paired-margin loss\\\bottomrule
\end{tabularx}\end{table}

\begin{figure}[tbp]\centering
\includegraphics[width=\linewidth]{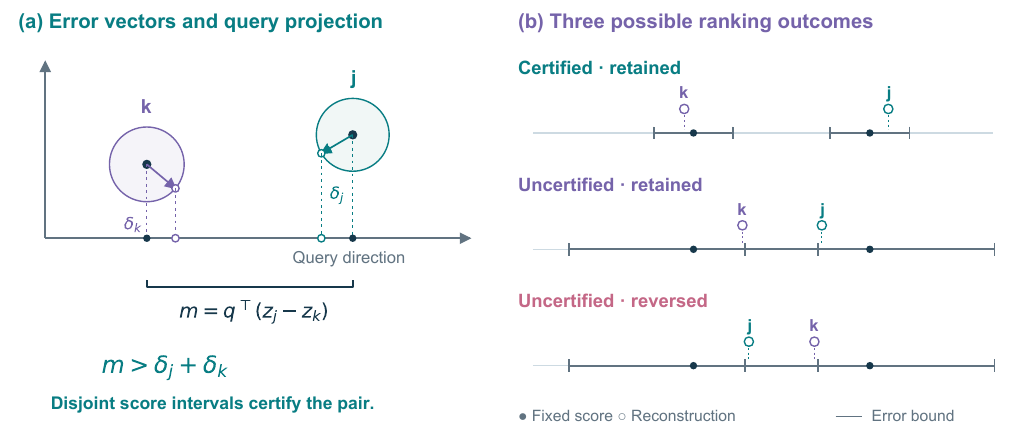}
\caption{Query-projected quantization errors determine a sufficient ordering certificate. Left, candidate reconstruction errors consume the fixed-provider score margin for a unit query. Right, separated error intervals certify retention; overlapping intervals permit either retention or reversal. Filled circles denote fixed scores and open circles reconstructed scores. Locations are schematic, not measured. Top-$K$ preservation requires the certificate for every boundary pair.}\label{fig:margin}
\end{figure}

\subsection{Optional upstream adaptation}

\begin{figure}[tbp]\centering
\includegraphics[width=\linewidth]{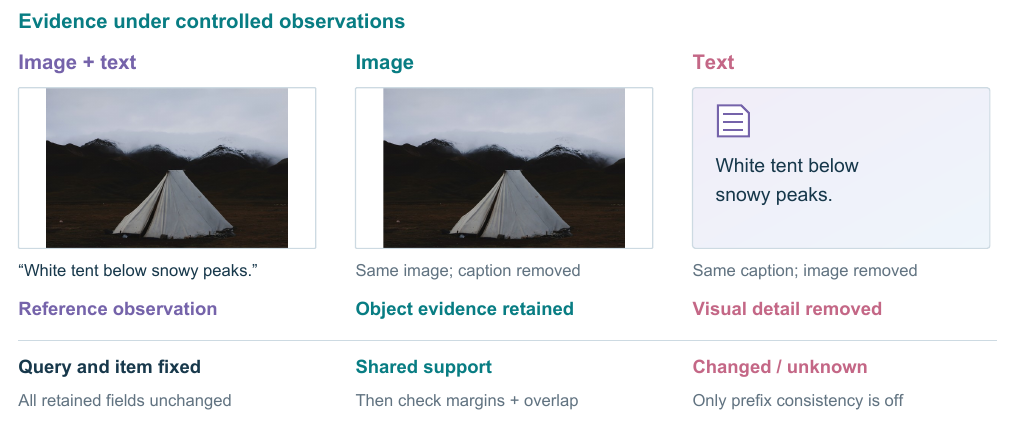}
\caption{An item-level evidence case connects retained observations to prefix supervision. The same tent image and authored caption define controlled views. Removing text preserves visible object evidence for witness review; removing the image eliminates entrance detail omitted by the caption. Audio is absent, so acoustic evidence remains unknown. Eligibility additionally requires independent evidence annotation, positive margins and overlap agreement. Changed or unknown evidence disables only the prefix term. This is a protocol illustration, not a model prediction. Photograph: adore chang, CC0.}\label{fig:partial}
\end{figure}

Figure~\ref{fig:partial} distinguishes observable content from a completed training label. Its visual query is an author-selected example, and the presence of a white tent can be checked in the pixels. The label needed by $\omega_{qjk}$ additionally requires a verified sufficient set, positive frozen-provider margins in both arms, and the overlap-agreement check. None has been executed for this illustration. A missing modality is therefore not a negative label, and visually plausible invariance is not a measured gate decision. For a detail that only the removed image supplied, the pair is a candidate for changed-evidence annotation; an acoustic query with no original audio remains unknown from the outset.

\subsection{Optional behavior adaptation}
A train-only sequence encoder may supply detached behavioral queries for a catalog-adapted extension. This extension receives its own checkpoint and result block, because MMQ, DIGER, and DIG already establish behavior-aware tokenization. The primary content tokenizer uses no consumer gradients and sets the behavior coefficient to zero. A leave-one-consumer-family-out experiment fits any behavior adapter without the held-out family's feedback and then trains that consumer on the exported codes.

\begin{table}[tbp]\centering\small
\caption{Registered downstream interfaces. MiniOneRec is an autoregressive instance. Dense retrieval has the measured outcomes in Table~\ref{tab:confirmed-main}; the new matched consumer-transfer outcomes are reported in Table~\ref{tab:new-consumers}.}\label{tab:consumers}
\begin{tabularx}{\linewidth}{p{.18\linewidth}YY}\toprule
Consumer family & Input and trainable interface & Shared-map evaluation\\\midrule
Dense / ANN retrieval & Reconstructed $\hat z$; frozen provider query & Recall, margin reversals, memory, query latency\\
Autoregressive SID & Level-specific symbols; fresh token embeddings and causal model; TIGER/MiniOneRec-style interface & Full-catalog HR/NDCG; invalid IDs; beam cost\\
Parallel SID & Position-typed code symbols and masks; fresh parallel decoder & Full-catalog HR/NDCG; refinement steps; invalid IDs\\
Discriminative ranking & Sum/concatenation of learned per-level code embeddings appended to matched features & NDCG or AUC under the declared ranking task; calibration\\
Slate / listwise scoring & Per-item code features and candidate list; fresh listwise scorer & NDCG, coverage, diversity; identical candidate sets\\\bottomrule
\end{tabularx}\end{table}

For a ranking or slate consumer, code embeddings are newly learned lookup tables, not copied $128$-dimensional quantizer centroids unless a separate initialization control requests that choice. A parallel consumer preserves level namespaces and the same collision policy as the autoregressive consumer. The dense lane uses reconstructed vectors and reports any retained full-precision reranking store separately. The pipeline covers these recommendation stages through defined adapters; no single decoder architecture is required by the tokenizer.

The primary portability experiment fixes the tokenizer, provider, and catalog manifest before fitting the consumers. The comparison is paired within each family: baseline and proposed tokenizers receive the same consumer architecture, data, initialization distribution, training budget, and evaluator. Absolute scores from different families are not pooled into a common metric. The summary reports per-family gains and a predeclared count of families with positive paired effects. A tokenizer is not selected on the held-out family in the leave-family-out study.

\subsection{MiniOneRec autoregressive integration example}
We audit the public MiniOneRec report and repository as separate artifacts \citep{minionerec,minicode}. The report identifier is arXiv:2510.24431. The separate ReRe paper uses arXiv:2510.12211 \citep{rere}. The source snapshot used for the contract is commit \texttt{526306229326013a7ca46f5b2d8f55ea669aa159}. The repository contains additional tokenizer alternatives and fixes beyond the original report. A reproduced baseline must identify both its code commit and the paper table it aims to reproduce.

The item index is a JSON mapping from string item IDs to ordered token strings. The inspected generator uses level-specific namespaces such as \texttt{<a\_17>}, \texttt{<b\_6>}, and \texttt{<c\_203>}. The SFT code collects tokens from this mapping, adds them to the tokenizer, and resizes the model embeddings. The interaction converter emits original item IDs alongside \texttt{history\_item\_sid} and \texttt{item\_sid}; this makes item-level auditing possible. The MiniOneRec adapter converts the shared integer manifest into this repository-specific contract.

The autoregressive comparison fixes Qwen2.5-0.5B-base as a proposed practical backbone, the prompt templates, history truncation, optimizer, SFT update count, token budget, and constrained decoding implementation. ``Fixed decoder'' means the architecture and training recipe are held fixed; each tokenizer still requires a separately trained decoder checkpoint because its token meanings differ. Freezing the numerical decoder weights across incompatible vocabularies would be an invalid comparison. Larger backbones are scaling controls after the main effect is established.

SFT-only evaluation is primary because it isolates the SID contribution with fewer optimization confounds. A secondary SFT-plus-RL experiment reuses the same reward functions, rollout budget, and reference-policy setup across tokenizers. It cannot be merged with the primary table. No user preference summaries are generated through external APIs in this work. Alignment tasks use original item titles and descriptions where available; a task requiring unavailable pseudo-labels is disabled consistently across variants and documented as a deviation from the original report.

The repository notes historical issues with SID alignment leakage and constrained decoding behavior. We do not assume that a stated fix guarantees a clean experiment. Before training, an input/target audit verifies that held-out next-item strings are absent from prompt-side auxiliary fields. Before scoring, the evaluator checks the rate of invalid item strings and round-trips every exported SID. If constrained generation produces an invalid item, it remains an error in the denominator and is not repaired by selecting the nearest valid embedding after evaluation.

\subsection{Audio retrieval and availability}
AudioCaps' official repository lists $975$ test audio clips, whereas the ImageBind paper reports an evaluation set of $796$ clips. These are different evaluation universes, even if both are called AudioCaps \citep{audiocaps,imagebind}. Our benchmark cannot adopt the larger denominator for one method and the smaller one for another. We create a current available-media manifest, document the exclusions, and rerun every baseline on that intersection. The legacy $796$ result is displayed only as source-reported context. The five-caption structure is maintained, and each text query is paired with its source audio ID.

For Clotho, the original evaluation set contains $1{,}045$ clips, and later versions reorganize development and validation splits. We pin the dataset version and checksum every caption file. The evaluated candidate set contains one record per audio clip, regardless of the number of descriptions. For reverse retrieval, we state whether success requires any reference caption or evaluates each caption separately. The two directions do not share a denominator and are not averaged without an explicit macro definition.

Audio preprocessing uses the checkpoint's required sample rate and frontend. A proposed default caps the observed duration at $20$ seconds, but long clips are assessed with a deterministic multi-window policy as a sensitivity test. The same crop policy is used by all compared methods. Silent clips remain valid observations with an explicit silence indicator derived from the waveform; unavailable clips are missing. Time-shifted and background-noise corruptions are generated only for robustness analysis, and their random seeds and signal-to-noise settings are saved.

\subsection{Video retrieval and temporal evidence}
The MSR-VTT experiment uses the 1K-A test IDs and keeps its text-to-video direction separate from audio-plus-video variants. The number of frames, temporal sampling positions, image resolution, and whether soundtrack information is permitted are part of the protocol name. A visual-only baseline cannot be compared to an audio-video model as if the inputs were equal. We therefore run matched visual-only, audio-only, and combined-input conditions, and report the marginal value of each.

DiDeMo is evaluated with paragraph-level text aggregation and whole-video candidate identity. Concatenating captions, averaging caption embeddings, and treating each localized sentence as an independent query are different tasks. The main manifest records the chosen published paragraph construction. Temporal sensitivity tests shuffle frame order, repeat a static frame, and remove the final event. Their purpose is to determine whether improvement uses motion and event order or merely a better image encoder. These controlled interventions do not substitute for the original benchmark result.

Both video datasets originate from web media and may overlap with large pretraining corpora. Source IDs are matched against all new training examples, and exact/near-duplicate content checks are performed where media is available. The inherited checkpoint's full pretraining set may not be recoverable. In that case we state the unresolved exposure and use equal initialization for the causal comparison. Equal exposure controls a treatment contrast; it does not prove an absolute zero-contamination benchmark.

\subsection{Recommendation with genuine raw modalities}
MicroLens is selected because its original release provides user interactions together with titles, cover images, audio, and full videos \citep{microlens}. We start with the public 100K-user release and use the larger release only after the controlled study is stable. The item count and interaction count are computed from the downloaded original files, rather than copied from a differently filtered secondary benchmark. Items without a usable raw modality remain in an availability stratum if the task supports them; they are not replaced by features from an unrelated collection.

Two recommendation split regimes answer different questions. The main temporal regime prevents later events from informing earlier recommendations, including when those later events belong to other users. If reliable global timestamps are absent in a release, the experiment is labeled an ordered-history protocol and no claim about a global temporal cutoff is made. A second user-disjoint regime tests transfer to histories from unseen users while keeping item content visible. Cold-item evaluation hides selected items' training interactions but permits their static content at SID export; a stricter inductive regime withholds their content from quantizer fitting as well.

Amazon experiments are image--text transfer controls across the registered consumer families; the pinned MiniOneRec branch supplies a reproducibility anchor for the autoregressive lane. We explicitly rebuild the Industrial and Office subsets from a pinned Amazon2018 release using our declared filter and time split. The original MiniOneRec appendix mentions Office and Industrial but also contains a Toys-specific date window and inconsistent dataset references. Consequently, its source-reported scores are not treated as our exact baseline until the original artifacts can be reconciled. Our own matched baseline is regenerated from the same files used by \method. Missing audio/video in Amazon is a real modality limitation, not an invitation to synthesize substitutes.

\begin{figure}[tbp]\centering
\includegraphics[width=\linewidth]{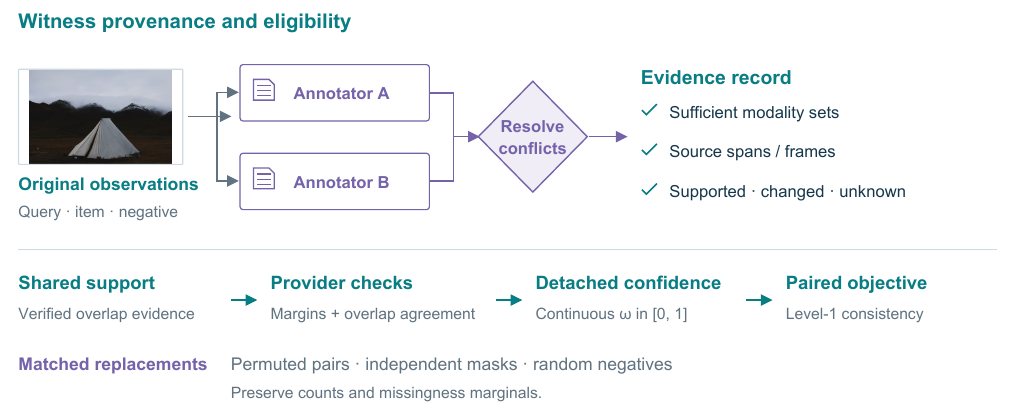}
\caption{Witness collection separates annotation from model-based eligibility. Independent annotators record sufficient modality sets and source locations, adjudicate disagreements, and then apply frozen-provider checks. Eligible observation pairs retain the same query and negative. Matched replacements remove pair coupling while preserving counts and missingness marginals. These fields specify the collection protocol and are not native public-benchmark annotations.}\label{fig:protocol}
\end{figure}

\begin{table}[tbp]
\centering\small
\caption{Source-reported LanguageBind video retrieval from arXiv v7 Tables10--13. R values are percent recall. All are downstream zero-shot, but backbone size and pretraining tuning differ. We retain the 3M LoRA block as the initial context for a parameter-efficient experiment.}
\label{tab:video}
\begin{tabular}{llrrrr}
\toprule
Configuration & Dataset & R@1 & R@5 & R@10 & Source table\\
\midrule
Large, 3M, LoRA & MSR-VTT & \reported{42.6} & \reported{65.4} & \reported{75.5} & 10\\
Large, 3M, LoRA & DiDeMo & \reported{37.8} & \reported{63.2} & \reported{73.4} & 10\\
Large, 10M, full & MSR-VTT & \reported{42.8} & \reported{67.5} & \reported{76.0} & 12\\
Huge, 10M, full & MSR-VTT & \reported{44.8} & \reported{70.0} & \reported{78.7} & 13\\
\bottomrule
\end{tabular}
\end{table}

\begin{table}[tbp]
\centering\small
\caption{Two separate source-reported contexts. GENIUS uses universal M-BEIR retrieval, with R@5 in percent. MiniOneRec reports next-item metrics as fractions. Neither block is a result of \method. The MiniOneRec dataset ambiguity prevents direct reproduction claims from this table alone.}
\label{tab:discrete}
\begin{tabular}{llrr}
\toprule
Source & Setting & Metric1 & Metric2\\
\midrule
GENIUS & MSCOCO universal & T$\to$I R@5: \reported{55.1} & I$\to$T R@5: \reported{82.7}\\
GENIUS reranked & MSCOCO universal & T$\to$I R@5: \reported{68.0} & I$\to$T R@5: \reported{90.6}\\
\midrule
MiniOneRec & Industrial, source protocol & HR@10: \reported{0.1586} & NDCG@10: \reported{0.1167}\\
MiniOneRec & Office, source protocol & HR@10: \reported{0.1634} & NDCG@10: \reported{0.1242}\\
\bottomrule
\end{tabular}
\end{table}

The GENIUS difference between code generation and reranking motivates our separate search lanes. It does not quantify the loss caused by our shorter SID budget, since GENIUS uses a different code configuration and task. Likewise, MiniOneRec's published recommendation performance motivates a reproducible integration target but does not predict the performance of an audio/video catalog. We use the original tables for context and the registered matched runs for causal conclusions \citep{genius,minionerec}.

\begin{figure}[tbp]\centering
\includegraphics[width=\linewidth]{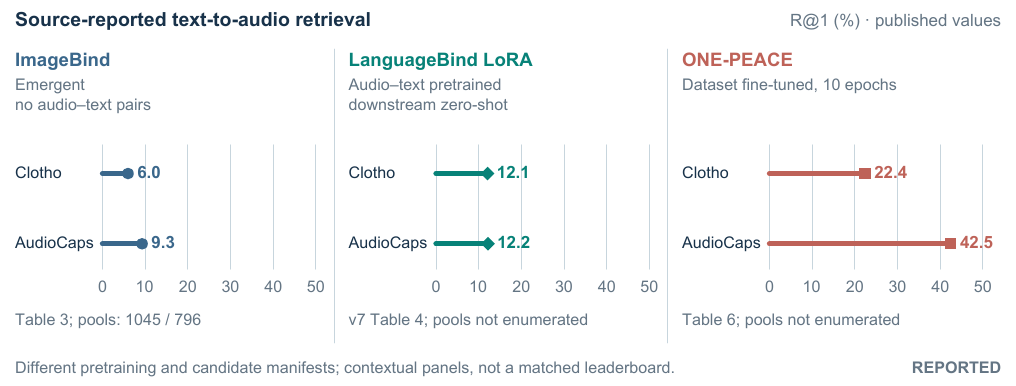}
\caption{Original-paper text-to-audio R@1. A common $0$--$50\%$ axis preserves numerical scale, while separate panels distinguish emergent alignment, audio--text pretraining, and dataset fine-tuning. Markers and direct labels reproduce the source values. Training access and candidate manifests differ; these contextual panels are not a matched leaderboard or results of \method. Source tables and pool information are recorded in Table~\ref{tab:audio} and the registry.}\label{fig:published}
\end{figure}

\begin{table}[tbp]
\centering\small
\caption{Prospective evaluation objectives (T). Target bands are design choices, not statistical intervals or measured scores. Acceptance conditions are applied to future matched experiments. CI denotes a paired $95\%$ confidence interval; improvement tests use the declared family-wise adjustment.}
\label{tab:targets}
\begin{tabularx}{\linewidth}{>{\raggedright\arraybackslash}p{.17\linewidth}>{\raggedright\arraybackslash}p{.18\linewidth}>{\raggedright\arraybackslash}p{.27\linewidth}Y}
\toprule
Condition & Metric & Target and acceptance threshold & Interpretation\\
\midrule
Adapted-provider retrieval & $\Delta$R@1, absolute pp & \target{+1\text{ to }+3} pp; corrected CI lower bound $>0$ & Optional provider-adaptation factorial only; frozen-provider continuous scores remain identical.\\
Quantized retention & R@1$_{\mathrm{quant}}$/R@1$_{\mathrm{cont}}$ & \target{0.95\text{ to }1.00}; CI lower bound $\geq0.95$ & The fixed three-code SID retains at least the declared fraction of continuous retrieval.\\
MicroLens recommendation & Relative NDCG@10 gain & \target{+3\%\text{ to }+8\%}; corrected CI lower bound $>0$ & SIDs improve next-item ranking within a matched consumer recipe; the historical numerical range is an AR-lane planning target.\\
Amazon image/text transfer & Relative NDCG@10 gain & \target{+3\%\text{ to }+8\%}; positive corrected effect in both domains & The effect transfers without invented audio or video observations.\\
Retrieval after catalog adaptation & Absolute R@1 decrease & \target{\leq1} pp relative to the content checkpoint & Validation selection constrains behavioral adaptation; this tolerance is not statistical equivalence.\\
Missing-audio robustness & NDCG@10 retention (missing / complete) & \target{0.90\text{ to }1.00}; retention $\geq0.90$ & Removal of audio is evaluated on the same users under the declared catalog policy.\\
\bottomrule
\end{tabularx}
\end{table}

A model is selected on validation results using a declared constrained objective: maximize recommendation NDCG subject to a retrieval degradation tolerance of at most one absolute percentage point and the fixed token budget. This selection tolerance is a design threshold, not a claim of statistical equivalence. The final test evaluates a single selected checkpoint per seed. If no checkpoint satisfies the constraint, the result is failure of the joint objective, and the unconstrained Pareto points are still reported.

\subsection{Component isolation}
The ablation ladder starts from a frozen embedding plus an independently trained residual quantizer. Continued contrastive pretraining adds exactly the same content pairs and updates as the full method. Reconstruction-aware adaptation adds the tokenizer reconstruction gradient. B3 adds the rank-margin and quantized-ranking terms to B2. B4 adds partial-observation consistency to B3; B5 instead adds behavior adaptation to B3. B6 combines both additions and is the full joint-adaptation configuration in the legacy AR planning lane; V4 is the frozen-provider primary proposal. A full factorial subset is run for rank preservation and partial-view consistency because their interaction may be important. Sequentially adding components alone cannot identify such interaction.

\begin{table}[tbp]
\centering\small
\caption{Historical joint-adaptation ablation definitions; these B-series arms are distinct from the newly measured fixed-provider V-series. The same downstream recipe is retrained for every tokenizer. No directional hypothesis is entered as an observed score.}
\label{tab:ablations}
\begin{tabularx}{\linewidth}{lYYY}
\toprule
Variant & Change & Primary diagnostic & Alternative explanation tested\\
\midrule
B0 & Frozen initializer + RQ & Continuous/quantized baseline & Starting geometry\\
B1 & Matched contrastive continuation & Continuous relevance & More data or updates\\
B2 & Reconstruction adaptation & Distortion and rank retention & Generic quantization awareness\\
B3 & Margin and quantized rank losses & Boundary reversals & Query-sensitive preservation\\
B4 & Supported subset consistency & Missing-view prefix stability & Ordinary modality dropout\\
B5 & B3 + train-only behavior loss & NDCG and retrieval trade-off & Behavioral alignment alone\\
B6 & B3 + subset + behavior & Joint retrieval/recommendation & Combined effect\\
C1 & Geometry-only control & Occupancy and anisotropy & GeoSID/CARD-type geometry effect\\
C2 & Shared-specific control & Cross-modal coverage & MMQ-type fusion effect\\
\bottomrule
\end{tabularx}
\end{table}

\subsection{Numeric ablation planning targets}
Table~\ref{tab:ablation_targets} preserves the earlier joint-adaptation AR-lane target registry for all nine variants. It is not the v4 frozen-provider primary experiment; the v4 arms are defined in Table~\ref{tab:v4design} and measured in Table~\ref{tab:new-controls}. Define $R_v=M_v/M_{B0}$, where $M_v$ is clean full-catalog NDCG@10 on the primary chronological MicroLens-100K split and $M_{B0}$ is the future measured score of the frozen binding-embedding plus RQ baseline. Every row uses the same denominator, SID rate, and MiniOneRec SFT recipe. B0's ratio of one is a normalization identity, not a measured score; if its measured denominator is zero, ratios are undefined and absolute metrics must be reported. B1--B6 and C1--C2 receive the same content and optimization budget; B0 intentionally retains its frozen encoder.

\begin{table}[tbp]\centering\small
\caption{Prospective ablation NDCG@10 ratios (T) on MicroLens-100K. The common denominator is future measured B0. All bands are design bands, not statistical intervals; no standard deviation or absolute baseline score is invented.}
\label{tab:ablation_targets}
\begin{tabularx}{\linewidth}{lrrY}\toprule
Variant & Target ratio & Design band & Planning rationale\\\midrule
B0 & \target{1.00} & \target{[1.00,1.00]} & Reference identity; denominator unmeasured\\
B1 & \target{1.01} & \target{[0.98,1.04]} & Small adaptation gain; possible regression\\
B2 & \target{1.02} & \target{[0.99,1.05]} & Recover generic quantization loss\\
B3 & \target{1.04} & \target{[1.00,1.08]} & Useful rank-preservation increment\\
B4 & \target{1.05} & \target{[1.00,1.09]} & Modest clean benefit from view consistency\\
B5 & \target{1.06} & \target{[1.00,1.10]} & Recommendation utility from behavior\\
B6 & \target{1.07} & \target{[1.00,1.12]} & Joint operating point; gains not additive\\
C1 & \target{1.03} & \target{[0.99,1.07]} & Competitive geometry-only explanation\\
C2 & \target{1.04} & \target{[0.99,1.08]} & Competitive fusion-only explanation\\\bottomrule
\end{tabularx}
\end{table}

The centers encode expected direction and practical scale for experiment planning; their ordering is a hypothesis, not evidence that every added component helps. Broad overlapping bands admit null effects and strong controls. The joint target is not the sum of component gains. C1 and C2 start from B2 and replace the proposed ranking and subset objectives with their respective geometry or shared-specific treatment under the common budget. This comparison isolates competing explanations instead of borrowing published scores from incompatible tasks. These ratios also have a different denominator from Table~\ref{tab:targets}, whose main recommendation target compares against a text-only tokenizer. Meeting an ablation center alone does not establish the main claim: the retrieval noninferiority tolerance and uncertainty analysis still apply. The CSV preserves every target when measured scores are later added.

Hard negatives are controlled independently of loss choice. One regime uses the same saved negative IDs for every variant; another refreshes negatives from each model's own training embedding. The former identifies objective effects under a common difficulty distribution. The latter assesses practical training behavior but can confound the objective with a changing curriculum. Both are useful when labeled. The main claim must not rely solely on whichever regime yields the larger number.

\begin{table}[tbp]\centering\small
\caption{Frozen-provider experiment matrix. The matched controls are now measured in Table~\ref{tab:new-controls}; provider-specific retention remains a separate endpoint. The baseline and full tokenizer are crossed with all consumer families in Table~\ref{tab:consumers}; mechanism replacements first use identical retrieval manifests and the AR anchor.}\label{tab:v4design}
\begin{tabularx}{\linewidth}{p{.10\linewidth}YY}\toprule
Arm & Intervention & Effect isolated\\\midrule
V0 & Frozen provider, standard RQ & Common representation and code-rate reference\\
V1 & V0 + fixed query margin/rank terms & Score-sensitive allocation beyond reconstruction\\
V2 & V0 + evidence-gated prefix consistency & Partial-view regularization alone\\
V3 & V0 + both objectives, independent sampling & Objective interaction without paired witnesses\\
V4 & V3 + paired near-boundary/view sampling & Full proposed coupling\\
V5 & V4 with view-pair permutation & Within-tuple coupling at matched marginals\\
V6 & V4 with all-level, ungated consistency & Supported coarse invariance versus forced identity\\
V7 & Anisotropic VQ replacement & Established score-sensitive quantization\\
V8 & GeoSID / PSRQ / MMQ-style replacement & Geometry, prefix, and expert-bank alternatives\\
V9 & GENIUS-style modality-decoupled codes & Shared versus decoupled discrete structure\\
V10 & PRO-style prefix ranking distillation & Closest frozen-provider ranking-preservation objective\\
V11 & V10 + proposed paired-view sampler and gate & Increment beyond the closest objective\\
V12 & MHR-style compact binary code at matched rate & Rank-anchored alternative bottleneck\\\bottomrule
\end{tabularx}\end{table}

\begin{table}[tbp]\centering\small
\caption{Provider--tokenizer factorial and consumer transfer protocol. New outcomes are reported in Appendix~\ref{sec:new-experiments}.}\label{tab:factorialv4}
\begin{tabularx}{\linewidth}{p{.18\linewidth}YY}\toprule
Factor / control & Levels & Attribution\\\midrule
Provider & Frozen; adapted using matched paired content & Continuous-geometry effect\\
Tokenizer & Standard RQ; V4 & Bottleneck effect within each provider condition\\
Consumer & Dense; AR; parallel; ranking; slate & Portability of the same exported map\\
Held-out family & Excluded from tokenizer/adapter selection & Transfer beyond a selected decoder\\
Provider swap & Binding model; unified omni-modal model & Dependence on upstream architecture\\\bottomrule
\end{tabularx}\end{table}

\subsection{Architecture exploration}
The component ladder tests losses within a fixed network. Table~\ref{tab:architecture-alternatives} separately tests structural choices. These are planned experiments, with no preferred outcome entered as a result. The first pass varies one family around the initial network. The second evaluates fusion placement jointly with projection sharing, because token-level mixing can change whether modality-specific projections are needed. Model selection uses validation data and the retrieval constraints in Table~\ref{tab:targets}; test scores do not choose the architecture.

\begin{table}[tbp]
\centering\small
\caption{Planned upstream and tokenizer architecture alternatives. Fusion, projection, residual-branch, and encoder-adaptation rows concern the optional provider factor; quantizer-depth and code-allocation rows concern the primary SID factor. The original network-alternative matrix is retained; measured rate and residual-depth sweeps are separately identified in Appendix~\ref{sec:new-experiments}. Fixed settings identify the causal comparison; quantities that necessarily change are recorded instead of being described as equal.}
\label{tab:architecture-alternatives}
\begin{tabularx}{\linewidth}{>{\raggedright\arraybackslash}p{.17\linewidth}>{\raggedright\arraybackslash}p{.26\linewidth}Y}
\toprule
Structure & Initial choice and alternatives & Control and diagnostic\\
\midrule
Fusion placement & Late fusion of pooled views; early token-level fusion; uniform pooled mean & Use the same source tokens and encoder checkpoints. Early fusion uses two shared attention blocks before pooling, with a fixed maximum token budget. Report trainable parameters, FLOPs, and retrieval by modality pair.\\
Projection sharing & Independent modality projections; common-width shared refinement; modality-specific refinement & Native input widths $D_m$ are retained. Sharing applies only after native-width bridges reach $768$. Compare tied and untied refinement maps, then a capacity-matched shared MLP; report both counts.\\
Residual branch & Rank $16$; rank $8$ or $32$; shared branch only & Hold the gate inputs and objectives fixed. Measure branch-norm ratio, continuous retrieval, and hard-code retention.\\
Quantizer depth & $768\!\to512\!\to128$; direct linear map; three-layer MLP & Keep the $128$-dimensional latent and $3\times256$ rate. Match training examples and report additional capacity rather than attributing it to depth alone.\\
Code allocation & $(L,K)=(3,256)$; $(2,4096)$, $(4,64)$, $(6,16)$ & All have $L\log_2K=24$ raw bits. Codebook parameters, vocabulary size, collision risk, and autoregressive token count differ and are measured separately.\\
Encoder adaptation & Final block plus projection; projection only; full encoder & Same checkpoint and data per comparison. Report trainable parameter count, update budget, and optimizer-state memory.\\
\bottomrule
\end{tabularx}
\end{table}

For early fusion, each observed encoder supplies its token sequence to the native-width bridge, followed by modality-type embeddings and two $768$-wide transformer blocks before masked pooling. The initial comparison uses $12$ attention heads and a $3072$-wide feed-forward sublayer in these added blocks. Their weights are proposed trainable additions, not parts attributed to the pretrained initializer. A shared maximum of $32$ sampled tokens per observed modality bounds the fusion input; the token sampler and temporal positions are fixed across the corresponding early-fusion controls. The optional prepared multi-tower provider keeps checkpoint-native pooling. Both provider variants are fitted before the frozen-tokenizer comparison; this table describes upstream architecture controls, not additional trainable modules in the primary SID method. Consequently this comparison is an architectural treatment with a changed attention budget, and its measured resource cost must accompany any accuracy claim.

Projection sharing cannot be implemented by tying unequal $D_m\times768$ matrices. The dedicated sharing experiment first supplies every variant with the same modality-specific native-width bridges, then compares identity, one shared $768\times768$ refinement map, and four independent maps before normalization. The tied-versus-untied contrast changes parameter count by construction. A wider shared MLP is an additional capacity control; it does not make the original contrast an equal-parameter comparison. We report these variants alongside the simpler main projector rather than silently replacing it.

A constant raw bit budget does not imply constant serving cost. The four code-allocation choices require respectively $2$, $3$, $4$, or $6$ autoregressive item tokens and have different codebook and vocabulary sizes. Each choice must pass its own injective-export check before recommendation training. Results are grouped both by raw bit rate and by measured decoder latency. Collision-suffix runs remain outside the fixed-length group, even if their uncorrected three-code capacity looks comparable.

\subsection{Rate, capacity, and compute}
A three-level $256$-entry SID has $24$ raw index bits before storage framing, collision handling, and the decoder vocabulary representation. It is not a claim of $24$ bits of total system memory per item. We count the item map, codebooks, vector decoder, trie, and any retained continuous embeddings separately. If reranking stores the original vector, the apparent catalog compression must include it. Latency includes query encoding, code generation, catalog lookup, and optional reranking, reported both separately and end to end.

The rate sweep uses $L\in\{2,3,4,6\}$ at $K=256$, with optional codebook-size controls at a fixed total raw bit budget. The new 16--96-bit measured sweep is reported in Table~\ref{tab:new-rates}. Table~\ref{tab:targets} specifies the prospective metric thresholds and their interpretations. These thresholds do not predict where a model will lie on a rate--accuracy curve. Different code lengths change generation cost, so the fixed-length main comparison precedes any best-rate comparison. Equal model parameter counts without equal update budgets are insufficient to claim efficiency.

\subsection{Uncertainty, aggregation, and failure reporting}
The new results provide five paired-seed means and difference SDs. Reported intervals use a paired-$t$ calculation across seeds. The separately registered source-item bootstrap would preserve all captions of a clip in a resample; the recommendation bootstrap would resample users with their target events. Those query/user sampling intervals are not reconstructed from the seed summaries. The synthetic exploratory scans retain their original three-seed spread.

We predeclare a small primary endpoint family: quantized audio R@1, quantized video R@1, AR-anchor MicroLens NDCG@10, and quantized retention. The parallel, ranking, and slate families form a separately corrected portability endpoint family. Continuous retrieval changes belong only to the provider-adaptation block. The current numerical intervals are pointwise; Appendix~\ref{sec:new-experiments} additionally gives Holm-adjusted tests for explicitly defined control families. These do not substitute for an unexecuted joint source-item/bootstrap primary analysis. Secondary slices are exploratory unless registered separately. Means are not weighted by the number of captions across datasets. We show each dataset and a declared macro average. A model that improves one large dataset while regressing on every smaller one cannot hide that pattern inside a global mean.

Failure logs include invalid codes, collision groups, empty histories, missing modalities, corrupted media, out-of-memory events, and numerical divergence. Runs are excluded only by rules set before observing test performance. A failed run is not replaced by a more favorable seed. If a baseline cannot be reproduced within its stated tolerance, its result is marked unresolved and the analysis uses the matched implementation with a documented deviation. This discipline is necessary for interpreting small gains in a pipeline with many opportunities for accidental changes.

The five consumer families define the integration scope. Their adapters are implemented and contract-tested. The new record supplies independently trained AR, parallel, ranking, slate, dense-reconstruction, and RecIF transfer endpoints. Canonical SID collisions, catalog churn, and different sequence-model capacities can change the effective rate and cost. The study must report these effects per catalog version. Source licensing, available media, and pretraining exposure may restrict the fully matched benchmark subset; the final submission will identify its exact evaluable population.

\section{Reproducibility Specification}
\subsection{Artifact organization and evidence states}
The source package contains the manuscript, bibliography, editable figures, a source registry, the historical experiment registries, and the current 17 November workbook. The leaderboard file stores source-reported values with metric, cutoff, direction, dataset edition, split, candidate pool, training regime, and evidence location. The current workbook is the submission result source for the 382 Omni-SID metric rows, including paired-seed summaries, consumer transfer, robustness, and rate--cost measurements. The historical CSVs and original target files remain archived for provenance; their rows are not additional independent training runs.

Evidence states identify provenance rather than rounding conventions. \texttt{REPORTED} denotes an original-paper value; \texttt{TARGET} denotes a predeclared numerical goal; and \texttt{PLANNED} denotes a configuration awaiting an outcome. The 17 November workbook is the current measured-results source for this submission; its record IDs and protocol keys are retained in the accompanying result export. Empty rows outside that workbook remain pending. The audit record preserves the provenance of each supplied value, while the recommended manifest below specifies the additional artifacts for independent reproduction.

The source snapshots are for audit, not redistribution of original datasets. Their checksums record which version was read. An original source can contain an error or an incomplete protocol; a checksum does not certify its truth. The source registry therefore includes unresolved ambiguities. In particular, MiniOneRec's report and current repository are distinct versions, and LanguageBind's initial arXiv versions differ from its final v7 tables. The manuscript uses v7 for the cited LanguageBind results and never combines a v3 score with a v7 configuration description.

\begin{table}[tbp]
\centering\small
\caption{Run manifest for independent reproduction. The measurement ledger records scores and source-protocol metadata; absent per-run fields remain unresolved. Historical target files remain unchanged; the new experimental CSV supplies the measured control outcomes.}
\label{tab:manifest}
\begin{tabularx}{\linewidth}{lYY}
\toprule
Group & Required fields & Why the field changes interpretation\\
\midrule
Identity & Run ID, commit, environment hash, timestamp & Reconstruct exact implementation\\
Data & Source release, hashes, split, exclusions, available views & Defines labels and denominators\\
Encoder & Checkpoint, trainable blocks, preprocessing & Controls inherited knowledge and capacity\\
Tokenizer & Dimension, codebooks, seed, collision policy, version & Defines the action space and rate\\
Optimization & Updates, effective batch, losses, schedule, precision & Controls training budget\\
Decoder & Backbone, vocabulary, prompts, SFT/RL recipe & Isolates representation treatment\\
Evaluation & Candidate list, masking, beam width, rerank flag & Defines ranking task and search cost\\
Hardware & Device, memory, batch size, warmup, synchronization & Makes latency interpretable\\
Evidence & Logs, checkpoint, predictions, bootstrap seed & Supports measured claims\\
\bottomrule
\end{tabularx}
\end{table}

\subsection{Proposed initial configuration}
The initial configuration is deliberately moderate: a public binding initializer, $768$-dimensional normalized projections, a $128$-dimensional quantizer latent, three codebooks of size $256$, and a Qwen2.5-0.5B-base AR anchor plus the other registered consumer families. These settings permit a controlled experiment before a costly scale sweep. We do not state a GPU-hour budget without knowing the exact decoded-media throughput and hardware. Instead, every comparison matches the number of optimizer updates and effective paired examples, while reporting actual wall-clock time after execution.

\begin{table}[tbp]
\centering\small
\caption{Prospective configuration registry. Values are design choices, not optimized settings or measured performance. The full sweep is selected using validation data only.}
\label{tab:hyper}
\begin{tabularx}{\linewidth}{lYY}
\toprule
Component & Initial setting & Controlled alternatives\\
\midrule
Shared dimension & $768$ & $256,512,1024$\\
Quantizer latent & $128$ & $64,256$\\
Residual codebooks & $L=3,K=256$ & $L=2,4,6$; fixed-bit alternatives\\
Shared prefix depth & $p=1$ & $p=0,2,L$\\
Primary encoder scope & Frozen provider, including projections and fusion & Adapted provider in separate factorial\\
Missing-view sampling & Paired near-boundary witnesses and supported/changed strata & Independent subsets; tuple permutation\\
AR anchor decoder & Qwen2.5-0.5B-base & Parallel, ranking, and slate families separately\\
Primary training regime & SFT only & Matched SFT plus RL secondary\\
Evaluation beam width & Fixed per experiment, initially $20$ & $10,50,100$ with latency reporting\\
Primary random seeds & Five preregistered seeds & No favorable-seed replacement\\
\bottomrule
\end{tabularx}
\end{table}

Table~\ref{tab:network-shapes} preserves analytical counts for the optional prepared multi-tower provider, the tokenizer, and the AR adapter. The first five rows belong to the upstream provider and are frozen in the primary v4 experiment. Only the quantizer encoder, codebooks, and reconstruction decoder contribute to the primary tokenizer count of $1{,}017{,}728$. They exclude the public encoder bodies, task embeddings whose count depends on the task registry, and the downstream decoder. They are neither profiler output nor measured memory. Affine layers in $E_Q$, $D_Q$, and the gates include biases; the two fusion output matrices and low-rank factors do not. LayerNorm has learned scale and offset. Embedding and output-head rows are counted once when tied.

\begin{table}[tbp]
\centering\small
\caption{Proposed tensor interfaces and analytical parameter counts. $D_m$ comes from the pinned modality checkpoint; $V_{\mathrm{sid}}\leq768$ is the number of new used SID symbols. These are configuration facts and formulas, not execution measurements.}
\label{tab:network-shapes}
\begin{tabularx}{\linewidth}{p{.25\linewidth}p{.30\linewidth}Y}
\toprule
Module & Tensor interface & Added parameter count\\
\midrule
One modality projection & $B\times D_m\to B\times768$ & $768D_m+2304$ including LayerNorm\\
Two fusion gates & Per observed view: $774\to64\to1$ & $99{,}330$ total, shared across modalities\\
Four rank-16 adapters & Four factors: $768\to16\to768$ & $98{,}304$ total\\
Fusion output matrices & Two maps: $768\to768$ & $1{,}179{,}648$ total\\
Learned residual scale & $\alpha=\operatorname{sigmoid}(a_\alpha)$ & $1$ scalar\\
Quantizer encoder $E_Q$ & $B\times768\to B\times128$ & $459{,}392$\\
Residual codebooks & $3\times256\times128$ & $98{,}304$\\
Reconstruction $D_Q$ & $B\times128\to B\times768$ & $460{,}032$\\
Added tied SID embeddings & $V_{\mathrm{sid}}\times896$ & $896V_{\mathrm{sid}}$; at most $688{,}128$\\
\bottomrule
\end{tabularx}
\end{table}

The vocabulary manifest records original tokenizer size, final resized model vocabulary, and the actual added-token list separately. A model configuration's padded vocabulary size is not used as a substitute for the tokenizer length. Reserving all $768$ namespace symbols rather than only used symbols is a separate export option and must be held fixed across compared tokenizers. All added embeddings are initialized by the pinned library routine before SFT; no continuous codeword vector is copied into a $896$-wide decoder row without a separately declared projection. This prevents an unreported initialization advantage from entering the representation comparison.

Loss coefficients are selected with a limited validation search. Rather than giving arbitrary magnitudes equal authority, each loss is first normalized by its detached running scale during a short training-only calibration window. The normalized coefficients are then searched over a small log grid. The calibration statistics are reset for each seed but use the same sample IDs. This procedure is an implementation option that must itself be held fixed across variants; a raw-weight alternative tests whether conclusions depend on normalization.

The main optimizer is AdamW with a shared learning-rate schedule for matched variants. In the provider-adaptation factorial, encoder adapters and codebooks may use different rates, but those rates are not retuned independently for each ablation after test inspection. Gradient clipping uses the same threshold and reports the fraction of clipped steps. A model that clips far more often may have an effectively different optimization regime, so clipping statistics belong in the reproducibility table rather than only the training log.

\subsection{Residual assignments, gradients, and export}\label{sec:tokenizer-implementation}
Figure~\ref{fig:network} gives the forward graph. This subsection fixes an initial implementation convention for the otherwise standard straight-through residual quantizer. It is a proposed specification, not a description of a completed run. For a batch of $B$ provider vectors, $Z\in\mathbb R^{B\times768}$ is detached, $U=E_Q(Z)\in\mathbb R^{B\times128}$, each residual tensor has shape $B\times128$, and each level returns $B$ integer assignments plus a $B\times256$ soft-assignment tensor. All three item/view branches reuse the same module objects and codebooks; they are not three separately fitted tokenizers. Queries are scored in the provider space and do not pass through $E_Q$ in the primary candidate-only compression experiment.

Write $v_\ell=e^{(\ell)}_{c_\ell}$ and $\tilde u=\sum_\ell v_\ell$. A hard forward pass chooses nearest entries with a deterministic smallest-index tie rule. For reconstruction and score losses, an initial straight-through convention uses
\begin{equation}
 u_{\mathrm{ST}}=u+\sg(\tilde u-u),\qquad
 \hat z=\operatorname{normalize}(D_Q(u_{\mathrm{ST}})).
\end{equation}
Its forward value equals the hard codeword sum, while reconstruction, geometric-margin, and score-ranking gradients reach $E_Q$ through the identity derivative with respect to $u$ and update $D_Q$. These three terms do not directly differentiate into the selected codewords. The selected codewords receive their direct updates through codebook terms and the soft-assignment objectives. In this convention, the residual update is evaluated as $r^{(\ell+1)}=r^{(\ell)}-\sg(v_\ell)$ for differentiation, with the same numerical residual as the hard forward recursion. Nearest-index selection itself is not differentiated. An EMA-updated codebook would be a different implementation convention and must be declared and matched across replacements rather than mixed into only one arm.

The initial reconstruction/codebook objective is
\begin{equation}
\begin{aligned}
 \calL_{\mathrm{rq}}=\E\Biggl[\|z-\hat z\|_2^2+
 \sum_{\ell=1}^{3}\Bigl(&\|\sg(r^{(\ell)})-v_\ell\|_2^2\\
 &+\beta\|r^{(\ell)}-\sg(v_\ell)\|_2^2\Bigr)\Biggr].
\end{aligned}
\end{equation}
The codebook term moves a selected vector toward its detached residual; the commitment term moves the encoder-facing residual toward the selected vector. The coefficient $\beta$ is a validation-selected design parameter shared by compared RQ implementations. Subset consistency and occupancy use the distance-based soft probabilities defined below. They do not replace the hard residual in the next stage. Numerical stabilization uses an ordinary log-sum-exp calculation for the soft distributions and a declared positive denominator floor for unit normalization. The floor and tie rule are part of the export manifest.

During the first $10\%$ of updates, reconstruction/codebook fitting, commitment, and occupancy initialize the bottleneck without query-pair or subset terms. The next $70\%$ enables the declared paired losses and temperature decay. The final $20\%$ keeps the same hard-forward graph and all enabled objectives, holds the temperature at its terminal value, and disables any training-only dead-code reinitialization. Thus ``hard-code stabilization'' does not mean that training previously used soft residuals or that gradients disappear in the final phase. The schedule is an initial proposal; validation sensitivity varies stage durations at fixed total updates. Initialization and any dead-code policy are identical for matched quantizers.

At export, $E_Q$, $D_Q$, and the codebooks are frozen. Each item's canonical observation policy produces exactly one provider vector and one hard tuple. Subset tuples used during training are not silently added as alternative retrieval documents. The raw tuple namespace contains $256^3$ possible strings, but this capacity does not imply collision-free assignment. The inverse map must resolve every exported string to a unique item under the declared collision policy, or the fixed-length result is marked infeasible. Storing the three indices as bytes uses three bytes per raw tuple; the inverse map, model parameters, alignment padding, and optional suffix consume additional memory.

The same release is then adapted independently to each consumer. Dense retrieval decodes codeword sums through $D_Q$ and retains the exact provider query. An AR consumer maps $(\ell,c_\ell)$ to a typed token and trains its own embedding rows; a parallel consumer additionally needs a declared mask symbol and level positions. A ranker uses level-specific lookup features with a matched scoring backbone. A slate model uses these item features together with a fixed candidate list. None of these adapters changes the exported assignments in the primary study. Consequently, code errors and consumer search errors can be measured separately, and replacing a tokenizer requires a new consumer fit under the same initialization and training recipe.

\subsection{A loss implementation that preserves masks}
Every batch contains explicit boolean masks for modality presence, positive relations, candidate eligibility, and duplicate membership. Masking occurs before normalization where appropriate. An absent view is not passed through its encoder and does not create a zero vector in the attention softmax. A query with no usable view is an input error and is counted. A candidate with no usable view can remain in a recommendation catalog only if the baseline and method share a documented fallback; it cannot receive a privileged learned embedding available to only one variant.

For contrastive losses, a row with no positive after filtering is skipped from that loss and counted in a coverage statistic. It is not assigned an arbitrary self-positive. For ranking losses, negatives are drawn from items eligible at the query timestamp, with all known positives removed. False negatives remain possible in implicit-feedback data, so the behavioral term uses a bounded weight and is not interpreted as a complete preference ordering. We report how often a nominal negative later becomes a positive in held-out data as a diagnostic computed only after training.

The subset consistency weight is computed from available overlap features and detached. At least one overlap view is required. We separately log the coverage of the consistency objective, because a high apparent robustness score could result from applying the objective only to easy examples. Coverage is stratified by modality mask and content duration. The main experiment includes all eligible test items regardless of their training consistency weight.

\section{Detailed Mathematical Analysis}
\subsection{Query quantization}
The main propositions keep the query exact. If the query is also quantized, let $e_q=\hat q-q$, $e_j=\hat z_j-z_j$, and $e_k=\hat z_k-z_k$. The perturbed pairwise gap satisfies
\begin{align}
\hat q^\top(\hat z_j-\hat z_k)-q^\top(z_j-z_k)
={}&q^\top(e_j-e_k)+e_q^\top(z_j-z_k)\\
&+e_q^\top(e_j-e_k).
\end{align}
For unit $q,z_j,z_k$, its absolute value is bounded by
\begin{equation}
 B_{qjk}=\delta_j+\delta_k+2\delta_q+\delta_q(\delta_j+\delta_k).
\end{equation}
A continuous gap larger than $B_{qjk}$ therefore preserves its sign. This bound uses the norm of the unquantized candidates and does not assume the quantized query is unit length. If reconstructed vectors are renormalized, the errors must be computed after that operation. Applying a bound to pre-normalization error while evaluating post-normalization scores would be inconsistent.

The additional query term explains why candidate-only and two-sided compression are different experiments. Query quantization may save cache or transmission cost, but it can hurt every candidate comparison simultaneously. The main study avoids this confound. A later rate-allocation study can distribute bits between query and catalog vectors, using the same explicit perturbation accounting. No property of residual quantization alone eliminates the query error.

\subsection{A relevance-aware failure upper bound}
For a query with one relevant item $j^+$, define the continuous rank $r_c$ and quantized rank $r_q$. Let $U$ be the set of candidates that the positive outranks continuously but for which the sufficient preservation condition fails. Candidates already above the positive are at most $r_c-1$, and only candidates in $U$ can newly overtake it. Thus
\begin{equation}
 r_q\leq r_c+|U|.
\end{equation}
This is a worst-case upper bound, not a prediction of the realized rank. It can be loose because many uncertified comparisons remain correct. The bound is nevertheless useful for explaining why protecting a handful of near-boundary candidates can matter more than reducing average distortion over easy negatives. For multiple relevant candidates, the same statement can be applied to each relevant item, but a recall bound requires the task's exact definition of success.

The proof is combinatorial once the pairwise proposition is established. Every newly higher-ranked candidate must have reversed its continuous relation to the positive. Certified comparisons cannot reverse, so such a candidate must belong to $U$. Counting the originally higher-ranked candidates and all possible new reversals gives the inequality. Ties require a deterministic tie-breaking rule and a non-strict version with explicit equality handling; we use strict comparisons in the theorem and report tied-score counts separately.

\subsection{An impossibility example for missing views}
Suppose two items have identical visible frames and titles but different audio events. A text query asks for an alarm. With audio present, the items can be separated. With audio absent, any deterministic function of the remaining identical observations returns the same representation. No loss can make this representation identify the correct item for that query without additional information. Therefore exact fine-grained SID invariance across full and audio-missing views cannot be universally compatible with retrieval correctness.

This example motivates partial consistency rather than all-level agreement. A coarse prefix may still encode the common visible scene, while finer levels encode the alarm when observed. The missing-view representation should express uncertainty through its retrieval behavior and confidence, not manufacture the unobserved event. In a single-canonical-SID system, the online decoder still predicts catalog IDs built from the designated catalog observation policy. Re-encoding the same item under a missing view is a robustness analysis unless the catalog is intentionally rebuilt under that policy.

\subsection{Canonical IDs and alternate observations}
Section~\ref{sec:canonical-id} defines canonical export, fixed-catalog degradation, rebuilt-catalog evaluation, and the separate decoder-compatibility stress test. The downstream serialization and collision protocols below implement that interface.

\section{Data Construction and Leakage Audit}
\subsection{Static content versus interaction-conditioned text}
An item title published with the content is a static observation; a user's review written after consuming the item is not equivalent. In next-item prediction, using the target user's future review as item text exposes the answer and may encode preference labels. The Amazon adapter therefore permits original product metadata and images but excludes the evaluated target event's review text from encoder input. If aggregated reviews are studied, they are restricted to events before the global training cutoff and form a separate experiment.

The same principle applies to video comments, popularity counts, and engagement summaries. MicroLens provides rich associated information, but the primary four-modality representation uses only the original title, cover, audio, and video. Comments and likes are not silently added as a fifth modality. They can contain post-event information and would change both the content scope and leakage surface. A future metadata experiment must specify the observation timestamp of every field.

\subsection{Duplicate groups and temporal overlap}
A duplicate audit begins with exact source IDs and content hashes. For videos, the manifest stores source-video ID and start/end timestamps, so nonidentical clips from the same source can be flagged. For audio, waveform hashes after a canonical decode catch exact re-encodings poorly, so a perceptual fingerprint is used as an additional approximate signal. Images receive exact hashes and a declared perceptual similarity threshold. Near-duplicate detection produces groups for inspection; it does not automatically justify deleting difficult examples after seeing model errors.

Training and evaluation splits are created before fitting adapters, quantizers, or behavioral teachers. If a duplicate group spans splits, a fixed rule assigns the whole group to one split or excludes it from the new pretraining set. The number and nature of exclusions are reported. Caption paraphrases attached to the same media remain in the same source group during bootstrap resampling. A training caption that describes an evaluation clip is a leak even if its wording differs from every test query.

\subsection{Transductive and inductive tokenizer regimes}
In transductive catalog tokenization, the model may assign codes to all candidate items using static content, including items whose interactions are held out. It may not fit behavior losses with their future targets. Codebook fitting on all catalog content is a separate choice from merely encoding it with a fixed quantizer. Our primary content-transfer experiment fits the quantizer on training content and then encodes evaluation candidates without updating it. A catalog-transductive adaptation is an explicit secondary regime.

Cold-item recommendation has two nested definitions. Interaction-cold items have no training interactions but may have participated in content pretraining. Fully inductive items have neither interaction nor content exposure during fitting and are encoded only after the model is frozen. We report both when the data supports them. Calling the first setting fully cold-start would overstate its novelty. The source manifest records item exposure at the content, tokenizer, behavioral-teacher, and decoder stages separately.

\begin{table}[tbp]
\centering\small
\caption{Leakage threat model and concrete audit. These are required checks for future execution, not claims that all external pretraining corpora are fully observable.}
\label{tab:leakage}
\begin{tabularx}{\linewidth}{lYY}
\toprule
Threat & Check & Consequence if unresolved\\
\midrule
Same source media across splits & Source IDs, intervals, exact/perceptual hashes & Exclude from new training; state inherited exposure\\
Future review/comment in content & Field-level timestamp and provenance audit & Remove field or separate protocol\\
Target SID in auxiliary prompt & Token-level prompt/label boundary assertion & Reject training example and inspect generator\\
Held-out behavior in tokenizer & Training-index membership assertion & Invalidate affected checkpoint\\
Candidate-count drift & Shared manifest hash in every prediction file & Rerun all compared models\\
Collision hidden by remapping & Inverse-map cardinality and round trip & Reject fixed-length export\\
Test-guided checkpoint choice & Selection log with validation-only metrics & Treat test as development; obtain new holdout\\
\bottomrule
\end{tabularx}
\end{table}

\section{Downstream Adapter and Evaluation Details}
\subsection{SID serialization}
The export process writes one ordered token list per item. A human-readable string is formed by concatenating those tokens, but training consumes tokenizer token IDs and not character positions. The exporter verifies that each token string maps to exactly one added token and that concatenated SIDs decode to the same sequence. It also verifies that every history item and target item is present in the index. Unknown items are counted; they are not assigned a shared unknown SID that could accidentally receive credit for multiple targets.

Level namespaces are essential. If code index $17$ occurs at both the first and second residual level, the corresponding strings remain distinct. Vocabulary ordering is deterministic and saved alongside the model. Sorting token strings lexicographically can place index $100$ before index $11$; this is acceptable if the exact vocabulary is saved, but downstream code must not assume that token-ID differences equal code-index differences. The adapter therefore stores an explicit level/index/token-ID table.

The exported mapping and inverse mapping carry the quantizer checkpoint hash. A decoder checkpoint stores the same hash. Evaluation aborts on mismatch. This prevents a particularly damaging silent error: applying a decoder trained on one set of semantic code meanings to a newly regenerated map with the same vocabulary size. Dimensions matching is not sufficient for semantic compatibility.

\subsection{Collision handling and prefix diagnostics}
Full-code collision rate is the fraction of items sharing their complete code with at least one other item. Prefix collision is different: many items intentionally share an early prefix, and this can express useful coarse structure. We report prefix occupancy distributions, effective perplexity, maximum bucket size, and the fraction of semantically unrelated items within a bucket using held-out content labels. None of these statistics alone proves recommendation quality.

The fixed-length primary regime requires unique full codes after training. We do not repeatedly search random seeds until a collision-free mapping appears without reporting those attempts. If the method cannot meet the uniqueness requirement at the proposed rate, that is a feasibility result. A suffix regime can restore unique identities, but its extra bits and variable sequence lengths are part of the model. Suffix tokens must also be preserved in every auxiliary alignment task; truncating them would reintroduce collisions during training.

Prefix diagnostics are computed before any item-level popularity weighting and then repeated with interaction-frequency weights. The unweighted version describes the catalog, while the weighted version describes the observed traffic distribution. A tokenizer may look balanced by items but highly skewed by traffic. Both matter: the first affects code utilization and the second affects decoder training dynamics. The distinction is retained in the CSV schema and figure captions.

\subsection{Constrained decoding and catalog validity}
The decoder generates one next-item SID at a time. A trie restricts token continuations to valid catalog prefixes, with an end condition that corresponds to a complete item code. Beam search returns distinct candidate strings, and the evaluator deduplicates by item identity before computing top-$K$ metrics. Duplicate candidates consume generation effort and are reported; they do not create extra credit. If fewer than $K$ valid unique items remain, the missing slots count as misses.

The trie must represent the candidate catalog permitted by the evaluation protocol. For a temporal recommendation task, items unavailable at the prediction timestamp cannot appear merely because their content is present in the final dataset. Implementations can construct time-bucketed tries or mask completed candidates, but the chosen policy must be identical for all variants. The overhead of time-dependent eligibility belongs in latency reporting. A single global trie is a different assumption and should be labeled.

Seen-item masking is also explicit. Some tasks permit repeated consumption; others exclude prior history items. The main protocol adopts one rule based on the source task and reports it in every table. Switching from unmasked to masked evaluation changes the difficulty and can alter the apparent benefit of semantic tokenization. Invalid output strings, duplicates, and seen-item violations are separate counters, not merged into a generic error number.

\section{Statistical and Computational Analysis Plan}
\subsection{Paired estimands}
For retrieval, the per-source score is the mean of its caption-query indicators. The treatment effect is the mean paired difference between \method\ and its matched baseline over source units. For recommendation, the per-user score averages that user's held-out targets under the declared temporal policy. We bootstrap these units rather than individual tokens, candidate comparisons, or captions. The estimand is therefore clear even when some users or items have many more observations than others.

We distinguish absolute and relative effects. An R@1 change from $40$ to $42$ percent is two percentage points and five percent relative improvement. Tables state which is used. Recommendation fractions are not mixed with percentage-scale retrieval values in a shared axis. Retention ratios are accompanied by both numerator and denominator. This prevents an apparently high ratio caused by a weak continuous baseline from dominating model selection.

The planned primary analysis combines source-unit paired intervals and correction over its endpoint family. The current reported analysis uses observed paired-seed summary intervals, with explicitly labeled family-specific Holm calculations where supplied. Source-unit sampling uncertainty remains separate. If only one trained checkpoint is available, the paper can still report query-level uncertainty but must not claim stability across training randomness. We do not manufacture a standard deviation from repeated evaluation of the same deterministic checkpoint.

\subsection{Complexity and memory accounting}
Let $N$ be the number of catalog items, $d$ the continuous dimension, $L$ the SID length, $K$ the codebook size, and $B$ the beam width. Storing raw integer codes costs approximately $NL\lceil\log_2K\rceil$ bits before alignment and file-format overhead. The codebooks add $LKd_q$ floating-point values. The reconstruction network, item map, tokenizer vocabulary, and trie add separate costs. A continuous reranking cache adds $Nd$ floating-point values and can dominate the catalog memory.

An exhaustive continuous scorer costs $O(Nd)$ arithmetic per query after query encoding. A generated-ID path has a decoding cost that depends on the model, $L$, and $B$, plus catalog lookup and eligibility checks. It is not correct to call end-to-end generation constant time merely because trie traversal depends on code length: the chosen beam width, model attention, and preprocessing remain substantial. We report empirical latency only after synchronized measurement on specified hardware, including warmup and batch size.

Quantization-aware training adds both codebook operations and query--candidate losses. A full all-pairs margin term would be expensive and unnecessary; the proposed training samples supervised positives and a fixed number of admissible hard negatives. The exact all-catalog certificate is reserved for evaluation on manageable candidate sets. Large-catalog evaluation reports sampled-pair diagnostics unless a complete score pass is performed. Training sample complexity and evaluation certification coverage are distinct quantities.

\subsection{Compute-matched and quality-matched comparisons}
The main comparison is compute matched: same encoder initialization, training examples, optimizer updates, effective batch, and decoder recipe. A quality-matched comparison asks how much compute each method needs to reach a fixed validation threshold. These answer different questions. A slower method with higher asymptotic quality may win the first metric and lose the second. We register both axes but do not convert unmeasured step counts into wall-clock savings.

When modality encoders have different costs, matching the total number of items is insufficient. A batch of ten-second audio differs from a batch of one-minute video. We log decoded seconds, sampled frames, text tokens, and effective paired examples by modality. A modality-balanced schedule may require more decoding work than a naturally imbalanced corpus. The paper reports this cost instead of treating balancing as free.

\section{Robustness Matrix and Error Taxonomy}
\subsection{Missing, corrupted, and conflicting observations}
The robustness matrix crosses three interventions with each modality. Missing-view tests remove an observation and update the availability mask. Corruption tests retain the observation but degrade its signal under a known process. Conflict tests replace it with content from another item under a controlled semantic relationship. These are distinct failure modes: a robust missing-view gate may still be overconfident in a plausible but wrong title.

Conflicting views are sampled from the same broad category when available, so the test is not solved by detecting an obviously unrelated modality. For audio, we distinguish speech replacement, background soundtrack replacement, and time misalignment. For video, we distinguish frame-order shuffling from changing the scene entirely. For text, we distinguish removing a modifier from replacing the whole title. The corruptions are disclosed as synthetic robustness interventions; they do not become training data without a separate augmentation experiment.

\begin{table}[tbp]
\centering\small
\caption{Planned robustness matrix. The outcome variables are measured separately for continuous retrieval, quantized retention, and next-item recommendation.}
\label{tab:robust}
\begin{tabularx}{\linewidth}{lYYY}
\toprule
View & Missing & Corrupted & Conflicting\\
\midrule
Audio & No waveform; mask absent & Noise, clipping, time shift & Different soundtrack with matched category\\
Video & No frames; mask absent & Shuffle, repeat, drop final event & Different clip with similar cover\\
Image & No cover; mask absent & Crop, blur, color distortion & Promotional or mismatched cover\\
Text & No original title & Modifier deletion, truncation & Plausible but wrong title\\
\bottomrule
\end{tabularx}
\end{table}

\subsection{Qualitative analysis without cherry-picking}
Qualitative examples are sampled by a predefined rule from four groups: improved after quantization, degraded after quantization, improved recommendation with unchanged retrieval, and conflicting-modality failure. Each group includes frequent and rare items. The sampling seed and candidate list are saved. We display original permitted content where licenses allow, together with continuous and quantized neighbors and the generated SID. We do not fabricate thumbnails or audio descriptions to make a mechanism appear convincing.

Error labels include semantic mismatch, temporal-order confusion, acoustic-event confusion, code collision, coarse-prefix ambiguity, decoder search failure, and catalog-eligibility error. Multiple labels may apply. The taxonomy is descriptive until annotated by a documented process. In this draft no human annotation agreement is claimed. An eventual qualitative panel must identify whether its labels were assigned by authors, independent annotators, or deterministic rules.

\section{Complete Objective Definitions and Execution Sequence}
\subsection{A fully specified observation gate}
For the optional prepared multi-tower provider, the modality-fusion gate uses the following two-layer GELU network. Define $\phi_j^m\in\mathbb{R}^{774}$ as the concatenation of the normalized $768$-dimensional modality vector, a four-entry one-hot modality indicator, standardized log duration, and a duration-availability flag. With $W\in\mathbb{R}^{64\times774}$ and $v\in\mathbb{R}^{64}$, the score is $a_j^m=v^\top\operatorname{GELU}(W\phi_j^m+b)+b_o$, and the weight is
\begin{equation}
 w_j^{m,O}=\frac{\mathbb{1}[m\in O]\exp(a_j^m)}{\sum_{n\in O}\exp(a_j^n)}.
\end{equation}
Modality-availability masks are retained outside this feature vector, and missing views do not enter the sum. The duration-availability flag is zero for a non-temporal modality or an unavailable duration. Duration is standardized using training-only statistics, clipped to a preregistered range, and set to zero for modalities without temporal extent. The gate does not receive item ID, interaction count, user ID, target rank, or a retrieval-success label. These exclusions prevent an apparently content-based fusion layer from becoming an unreported popularity model.

The residual weights $\tilde w$ use the same availability mask and a separate small attention head. The residual scale is parameterized as $\alpha=\operatorname{sigmoid}(a_\alpha)$. This parameterization bounds the scalar coefficient, not the residual branch norm, because the projection matrices are unconstrained. We report its learned value, the ratio of residual to shared branch norms, and an ablation fixing it to zero. Both fusion output matrices start at the identity. The $U_m$ residual factors start from zero-mean Gaussian entries with standard deviation $10^{-3}$, while $V_m$ uses Xavier initialization. Native-width projections use the pinned framework initialization and are aligned during semantic warmup; they are not claimed to reproduce the initializer exactly when its width differs. These rules define an initially small residual path without preventing either factor from receiving a gradient.

The exact subset-confidence definition is given in Equation~\ref{eq:evidence-gate}. A sigmoid calibration is a separately registered alternative, not the implemented default. Before benchmark fitting, the threshold may be calibrated on a held-out training partition; the same policy must then remain fixed across test subsets. We report both supported-stratum coverage and confidence coverage $\omega\geq0.5$, stratified by observation mask. Calibration targets of $50\%$, $80\%$, and $100\%$ within eligible pairs are sensitivity settings rather than achieved coverage. Calibration on held-out test retrieval correctness is not permitted.

\subsection{Codebook distributions and usage regularization}
Equation~\ref{eq:soft-assignment} and the main optimization specification give the reference soft assignment, temperature schedule, and fixed occupancy target. A soft-residual alternative changes the training dynamics and is a separate ablation. A training-median-scaled initial temperature is another sensitivity control; the selected scale is saved because it depends on latent normalization. An online moving-average occupancy target is a separately registered alternative to the fixed target. Uniform usage, empirical usage, and no usage loss are separate controls. Code balance and relevant-ranking preservation remain distinct diagnostics.

In the optional provider-adaptation factor, the neighborhood teacher distribution is a softmax of frozen-initializer similarities over a saved training queue. Its temperature is calibrated to a fixed median entropy, and candidate IDs are shared across the student and teacher. We clip the per-query KL contribution at a fixed cap so a single incompatible teacher neighborhood cannot dominate a batch. The cap, excluded-positive policy, and queue refresh rate are registered before training. These details make the neighborhood anchor a reproducible regularizer rather than a vaguely defined distillation term.

\subsection{Two explicit content-training regimes}\label{sec:training-access}
The transfer-only tokenizer regime fits $E_Q,D_Q$, and the codebooks on fixed embeddings of original MicroLens training-item content and any additional public training corpus listed in a frozen manifest before the experiment. The minimal release uses MicroLens alone; it does not presume that a large external web corpus has been downloaded or licensed. Audio/video pairs come from the same original clip, cover images and titles retain their original item association, and no caption-generation API is used. Because a title need not describe a soundtrack faithfully, the loss treats same-item association as instance evidence and does not claim a human-verified acoustic caption. Pair-type weights and disagreement rates are reported.

In this transfer-only regime, AudioCaps, Clotho, MSR-VTT, and DiDeMo training labels are not used for adapter or quantizer fitting. Their validation subsets may select a shared hyperparameter configuration, in which case the result is downstream validation-selected transfer rather than completely untouched zero-shot evaluation. A stricter zero-shot variant selects hyperparameters solely on held-out MicroLens content and reports the external benchmarks once. The paper keeps these labels separate because validation exposure can materially affect model selection even without gradient updates.

The supervised retrieval tokenizer regime fits the tokenizer on fixed provider outputs from the original training splits of the corresponding retrieval datasets. It tests the attainable quality of the discrete bottleneck under direct task supervision and belongs beside other fine-tuned methods, not in a zero-shot block. A multi-task version balances these training splits by query units and excludes all evaluation media. Each fine-tuned checkpoint is evaluated before and after quantization on the same candidate manifest. This regime can answer whether ranking-preserving quantization helps specialist retrieval even if the transfer-only regime yields no improvement.

The recommendation study uses the same raw-content checkpoint to construct SIDs, then optionally applies train-only behavior adaptation. The four combinations of transfer-only versus retrieval-supervised content training and behavior-free versus behavior-adapted tokenization form a small, interpretable matrix. They are not pooled into one best model without disclosure. The primary recommendation comparison uses transfer-only tokenizer fitting with a frozen provider so that external retrieval supervision cannot silently explain its advantage over the matched baseline.

\subsection{Execution order and stop conditions}
Execution begins with manifest construction and source reconciliation. The next stage reproduces continuous retrieval from the chosen initializer without training, establishing that preprocessing and candidate handling are correct. It then trains a simple residual tokenizer and tests export round trips. Only after these checks pass are the proposed losses introduced. This sequence makes failures attributable: a broken decoder adapter cannot be mistaken for a weak representation, and a wrong caption denominator cannot be mistaken for a retrieval improvement.

The initial compute pilot runs only training and validation examples. Its purposes are numerical stability, memory estimation, and verifying that every loss receives gradients. Pilot curves are not reported as benchmark evidence. A loss whose gradient is always zero because of an accidental detach is a failed implementation; a loss with a large gradient but no validation benefit is a valid negative result. Both are logged differently. Test evaluation begins only after the objective weights, negative-mining policy, and checkpoint selection rule are frozen.

A run stops early for nonfinite loss, irrecoverable media corruption, or a violated data-access assertion. Stopping for poor test performance is prohibited. Validation-based early stopping uses the same patience and maximum updates across methods, while a fixed-update comparison remains the primary compute control. If a method fails to export a collision-free fixed-length codebook, its fixed-rate result is reported as infeasible and its suffix-enabled result is moved to the engineering regime. The next experiment is selected from the preregistered matrix rather than by inventing an explanation after every unfavorable outcome.

\section{Decision Rules for a Final Empirical Paper}
The primary empirical test requires reproducible baselines, identical continuous scores under a frozen provider, improved quantized retention at fixed SID budget, and utility gains within matched consumer families. Continuous retrieval improvement is an endpoint only of the optional provider-adaptation factor. Missing-view and geometry controls must explain the effect. Failure at any stage narrows the final claims while preserving earlier valid findings.

If the full joint objective fails, the most useful revision may be a retrieval-preserving tokenizer paper without a broad recommendation claim. If behavioral adaptation dominates, the paper should focus on the semantic--behavioral trade-off and report the retrieval cost. If PRO plus independently sampled missing views matches the proposed joint sampling rule, the proposed coupling is not supported and should not be retained through selective reporting. These decisions are recorded before the experiment to reduce pressure to reinterpret every outcome as success.

The historical fourteen tokenizer measurements and 168 baseline entries remain archived. The 17 November workbook supplies the current 382 metric rows for tokenizer scores, retention, consumer transfer, robustness, and rate--cost behavior. The resulting conclusions use these current contrasts while preserving the original registries as provenance records.

\section{V4 Novelty Audit and Experiment Registry}
The audit directory contains a query log, primary-source reading records, pinned-source checksums, and a three-axis comparison. The historical \texttt{data/novelty\_v4\_experiments.csv} remains a design registry; measured counterparts are now linked by record ID in \texttt{data/confirmed\_experiments\_20261112.csv}. The earlier main-results, leaderboard, and ablation CSV files are preserved byte-for-byte. Their continuous-improvement and adapted-encoder targets belong to the optional upstream/AR planning lane, as identified in the v4 registry; they do not supply predicted outcomes for the frozen-provider study.

The closest-method replacements are evaluated at equal effective catalog rate, not merely an equal nominal number of residual indices. For modality-specific banks, extra bank identifiers and any collision suffix count toward the rate. For reconstruction-based retrieval, retained vectors and decoder parameters count toward memory. For a learned consumer, code embeddings and model parameters count toward the separately reported training and serving costs. These choices prevent a multi-bank system from receiving extra hidden capacity or a reconstructed-vector system from receiving an unreported reranking advantage.

\subsection{Training tuple manifest}
Each tuple records task ID, source query ID, positive/negative item IDs, full and subset observation masks, frozen-provider hash, continuous scores, boundary-quantile bin, evidence stratum, sufficient modality-set list, original evidence locations, annotator/adjudication IDs, consistency weight, and sampling probability. An independent-sampling replacement preserves the marginal distribution of all masks and query families. The tuple-permutation replacement additionally preserves stratum counts while breaking the association between a particular near-boundary witness and its observation pair. This distinction is needed to attribute an effect to the joint sampling rule.

The evidence label is verified separately from provider scores. The supported indicator requires an annotated sufficient set $A\in\mathcal{A}_{qjk}$ with $A\subseteq O\cap O'$, both positive fixed-provider margins, and the fixed agreement threshold. A missing annotation is unknown, not implicitly supported. The reconstruction and rank terms use observable evidence. A query whose labeled distinction requires an absent modality is not assigned an artificial positive comparison in that arm. Such cases remain in an explicitly scored missing-evidence slice. The supported-evidence confidence is detached, and the final subset weight includes the indicator for the supported stratum. Coverage is reported alongside retrieval so that a gate that selects very few pairs cannot appear effective solely by avoiding difficult examples.

\subsection{Catalog release and consumer manifests}
The item manifest stores one canonical tuple per catalog item under the declared observation policy. The dense adapter decodes the selected vectors. The AR adapter converts each residual index to a level-specific token. The parallel adapter additionally exposes the level position and mask symbol. The ranker adapter embeds each level into its own learned table and concatenates or sums those features under a pinned choice. The slate adapter uses the same item features and a fixed input candidate set. All adapters point to the same item-ID inverse map, allowing evaluator-side round trips and full-catalog denominators.

A consumer run records the tokenizer release hash, provider hash, adapter version, model initialization, training split, candidate manifest, output item map, and optimizer budget. A change in output vocabulary triggers a fresh model fit even when architecture and seed are held fixed. A consumer selected after seeing its test result cannot be counted as evidence of out-of-family transfer. The held-out-family selection rule is fixed before any result enters the cross-family summary.

\section{Visual Evidence and Source Provenance}\label{sec:visual-provenance}
The tent photograph is an explanatory input, not a benchmark case or trained output. \emph{Tent (Unsplash)} by adore chang is CC0 1.0 from the pre-June-2017 Unsplash collection. Its \href{https://commons.wikimedia.org/wiki/File:Tent_(Unsplash).jpg}{source page}, original file and checksum accompany the source package. The image is displayed with proportional scaling. Earlier retired figures and their source licenses remain archived separately.

The caption ``A white tent below snow-capped mountains'' was written by the manuscript authors after viewing the photograph. It is not source-dataset annotations, generated benchmark labels, or claims about acoustic content. The controlled case removes an entire modality: it does not alter the retained photograph or rewrite the retained caption. Its three illustrative query intentions distinguish visually available object evidence, an image-only entrance detail, and an unobserved acoustic event. The example supplies no provider inference, human agreement statistic, gate value, nearest neighbor, or SID assignment. In particular, shared visible evidence is eligible for the proposed annotation procedure but is insufficient to certify the supported stratum by itself.

\section{Extended Method Specifications}\label{sec:extended-method}
The following specifications expand the main-text computation graph and objectives.
\subsection{Provider initialization}
For compatibility with a multi-tower initializer, an optional upstream preparation can form normalized views $h_j^m$, a masked weighted sum $s_j^O$, and low-rank residual features $r_j^O$. Its output is
\begin{equation}
 z_j^O=\operatorname{normalize}(W_s s_j^O+\alpha W_r r_j^O).
 \label{eq:fusion}
\end{equation}
This entire expression belongs to the provider. It is fitted once and frozen, or replaced by the native embedding output. Shared-specific fusion is an established alternative motivated by MMQ; its presence is not attributed to the proposed SID objective. A separate two-by-two experiment crosses frozen versus adapted providers with baseline versus proposed tokenizers.

\subsection{Typed tokenizer interfaces}
The tensor interfaces separate four objects that are easily conflated: the provider vector $z\in\mathbb R^{768}$, tokenizer latent $u\in\mathbb R^{128}$, sum of selected codeword vectors $\tilde u\in\mathbb R^{128}$, and integer tuple $c\in\{0,\ldots,255\}^3$. The reconstruction decoder receives $\tilde u$, never the three raw integers. A generative consumer instead receives three level-typed symbols; its embedding tables are learned during that consumer's own fit. This difference matters for attribution: a continuous codeword is not silently injected into MiniOneRec's $896$-dimensional token embedding. The nominal tuple carries $24$ bits before collision handling, storage alignment, or any suffix, which are accounted for separately.

\subsection{Witness annotation protocol}
The required witness annotation is an additional training-data task. Ordinary query--item relevance files do not contain it. For each $(q,j,k)$, the record stores a set $\mathcal{A}_{qjk}$ of sufficient modality subsets, original-source evidence locations (audio/video timestamps, frame IDs, or text spans), and supported/changed/unknown labels for the planned interventions. Two annotators independently inspect the original observations and candidate comparison; disagreement is adjudicated before the record can enter the supported stratum. If no sufficient set is verified, $\mathcal{A}_{qjk}=\varnothing$ and the record remains unknown. The release includes annotation agreement and adjudication counts. These are proposed collection steps, not annotations claimed to exist in MicroLens or the retrieval benchmarks. Cosine agreement supplies a confidence after this evidence check; it never creates the evidence label.

\subsection{Ranking preservation: proofs and diagnostics}
\begin{proposition}[Candidate quantization and an exact query]
Let $q$ be a unit vector, let $z_j,z_k$ be continuous candidate vectors, and let $\hat z_j,\hat z_k$ be their reconstructions. If
\begin{equation}
 q^\top(z_j-z_k)>\norm{z_j-\hat z_j}_2+\norm{z_k-\hat z_k}_2,
 \label{eq:certificate-supp}
\end{equation}
then $q^\top\hat z_j>q^\top\hat z_k$.
\end{proposition}
\begin{proof}
Write $e_j=\hat z_j-z_j$ and $e_k=\hat z_k-z_k$. The quantized gap equals $q^\top(z_j-z_k)+q^\top(e_j-e_k)$. Cauchy--Schwarz and the triangle inequality bound the second term below by $-\norm{e_j}_2-\norm{e_k}_2$. The assumed strict inequality makes the resulting lower bound positive.
\end{proof}

This elementary sufficient condition follows the decision-stability perspective of prior quantization analyses~\citep{decisionquant,retrievaldamage} and is independent of the training method. It is valid for any quantizer that returns the stated reconstructions. It does not imply that its sufficient condition is necessary: rankings may survive even when the bound is negative. The useful diagnostic is therefore a three-way partition into certified preserved pairs, uncertified but preserved pairs, and reversed pairs. Reporting only the first group would hide both conservatism and failure. The confidence interval for each group is computed over independent query units, with captions from the same source item grouped together.

\begin{proposition}[Top-$K$ stability from boundary comparisons]
For a fixed query with distinct continuous scores, let $S_K$ be its continuous top-$K$ set. If Equation~\ref{eq:certificate} holds for every $j\in S_K$ and $k\notin S_K$, the quantized top-$K$ set is exactly $S_K$.
\end{proposition}
\begin{proof}
Every item inside $S_K$ continues to outrank every item outside it. Consequently no outside item can enter the first $K$ positions. Internal order may change, but membership cannot.
\end{proof}

Checking all boundary pairs is expensive in a large catalog. Evaluation can use a sufficient aggregate test: the smallest continuous inside score minus its largest error budget must exceed the largest outside score plus its error budget. This bound is looser but requires only reductions over scored candidates. Sampling a few hard negatives does not certify the whole top-$K$ set, and the paper will label sampled-pair diagnostics accordingly. Approximate nearest-neighbor retrieval cannot supply an exact all-catalog certificate without an additional bound on omitted candidates.

The theorem speaks about preservation of a continuous ranking, not improvement in relevance. A bad continuous model can preserve its mistakes perfectly. For this reason, our primary endpoints require a fixed continuous relevance reference, high quantized retention, and consumer-level utility. We also evaluate the model's true relevant items directly, not merely its agreement with itself. The distinction prevents a trivial identity-like tokenizer for a weak representation from appearing scientifically successful.

\subsection{Three-axis design comparison}
Table~\ref{tab:threeaxes} separates an implementable difference from a demonstrated contribution. The training, architecture, and data axes are coupled through explicit quantities in Section~\ref{sec:witness}. The inherited RQ backbone, multi-consumer interface, and use of a frozen provider are controls and system decisions. The research question is whether evidence-conditioned comparison sampling changes which information survives the code bottleneck.

\begin{table}[tbp]\centering\small
\caption{Proposed differences across three axes. These are testable design differences from the closest reviewed formulations, not three independently established algorithmic novelties.}
\label{tab:threeaxes}
\begin{tabularx}{\linewidth}{p{.12\linewidth}YY}\toprule
Axis & Proposed specification & Mechanism replacement\\\midrule
Training & Fixed-provider pair margins plus shared-evidence prefix consistency & PRO prefix distillation; anisotropic VQ; each objective separately\\
Structure & Shared canonical RQ path with a detached evidence gate on first-level agreement & MMQ-style expert banks; GENIUS-style decoupled codes; ungated consistency\\
Data & Near-boundary query triples paired with same-item view subsets; supported/changed-evidence strata & Independent hard negatives and missing-view augmentation at the same sample count\\\bottomrule
\end{tabularx}\end{table}

The comparison also includes DRQ, which separates continuous reshaping from residual clustering, and AST, which constructs adaptive prefix-conditioned trees \citep{drq,ast}. Neither code-boundary robustness nor hierarchical semantics alone establishes originality here. The primary-source audit extends through 11 November 2026; the remaining test is the incremental effect of paired witnesses and their evidence gate.

\section{Reference Implementation and Additional Controls}\label{sec:reference-code}
\begin{table}[tbp]\centering\small
\caption{Additional registered controls. These are prospective experiments; no scores are assigned. Original result ledgers remain unchanged.}
\label{tab:additional-controls}
\begin{tabularx}{\linewidth}{p{.16\linewidth}YY}\toprule
Control & Held fixed and replaced & Decisive endpoint\\\midrule
DRQ replacement & Same provider; continuous shaping then residual clustering instead of joint RQ & Retention, realized rank reversals, code overlap, stored rate\\
AST replacement & Same provider; adaptive tree IDs with explicitly charged depth/capacity & Item-level utility and prefix occupancy at matched storage\\
PRO increment & Prefix distillation alone versus the same objective plus paired witnesses/gate & Added utility beyond the closest training objective\\
Annotation control & Same verified examples; confidence-only, unweighted verified, and shuffled-witness gates & False support, coverage, missing-view retention\\
Coupling interaction & Boundary/random comparison sampling crossed with paired/permuted observation assignments & Difference-in-differences across all four cells\\
M-BEIR block & Identical queries, local/global pool declaration, fixed encoder and decoder recipe & Code-only retrieval, reranking separately, no cross-pool ranking\\\bottomrule
\end{tabularx}
\end{table}

\subsection{Executable scope and gradient ownership}
The accompanying reference implementation separates a frozen provider wrapper, a residual tokenizer, a paired-witness objective, and catalog adapters. It implements the candidate-only compression setting: queries remain in the provider space, while candidate vectors pass through $E_Q$, hard residual assignments, and $D_Q$. The provider wrapper disables parameter gradients and training-mode dropout. A cached-vector run requires a declared provider version, observation policy, original source IDs, and split manifest. Cached vectors do not establish the provenance of raw media by themselves. Reference validation passes 23 mathematical, gradient, gating, and export-contract tests, a 40-update synthetic CPU smoke run, and independent multi-positive evaluation/export interface checks. These executions verify mechanics and are separate from benchmark measurements.

The forward and backward contracts are distinct. Each next residual subtracts the numerical selected codeword, but the subtraction detaches that codeword for differentiation. The final reconstruction uses $u+\sg(\tilde u-u)$, so reconstruction, margin, and score-ranking terms update the encoder and decoder without directly updating codebook entries. Selected codewords are fitted by the codebook loss; the commitment term updates the encoder-facing residual; soft-assignment consistency and occupancy update both the relevant residual and current codebook. A control that routes score gradients directly into codewords is a different estimator and receives a separately named configuration. It cannot be silently substituted for this implementation.

The implementation averages reconstruction, codebook, and commitment contributions across the reference, alternate, and negative branches, which reuse the same modules. Ranking and geometric losses average across the two positive arms and the full batch; ineligible rows contribute zero. Thus changing evidence coverage changes the effective amount of supervision rather than being hidden by renormalization over a shrinking eligible subset. Every comparison reports coverage alongside its objective coefficient. The geometric term excludes nonpositive provider gaps because those comparisons cannot satisfy its preservation condition. The direct ranking term may still correct an annotated positive that the provider ranks incorrectly. Even a positive gap smaller than $\gamma$ need not admit zero geometric loss; training the hinge does not guarantee certification.

\subsection{Explicit evidence-gate contract}
Let $A$ denote the event that a verified sufficient set $S\in\mathcal A_{qjk}$ lies inside $O\cap O'$. Define the detached overlap agreement and a fixed calibration by
\begin{align}
 a&=\min\{\cos(z_j^O,\bar h),\cos(z_j^{O'},\bar h)\},\\
 c(a)&=\left[\frac{a-a_0}{1-a_0}\right]_0^1,\qquad -1\leq a_0<1,\\
 \omega_{qjk}&=\mathbf1_A\mathbf1_{m_O>0}\mathbf1_{m_{O'}>0}
 \mathbf1_{a>a_0}\,c(a).
\end{align}
Here $\bar h=F(x_j^{O\cap O'})$ uses the same frozen provider policy. An empty overlap sets $\omega=0$ without encoding invented observations. The reference configuration uses $a_0=0.5$ as an initial choice, not as a calibrated empirical result. Annotation verification precedes this calculation. If no sufficient set is recorded, even identical vectors have zero consistency weight. Conversely, a supported set in the alternate arm permits its ranking term even when no sufficient set is shared across both arms. These are separate eligibility decisions.

The cache schema stores observed masks, verified sufficient sets, original comparison relevance, overlap vectors, and ambiguity weights. Human records retain evidence locations and adjudication metadata outside the tensor cache. The implementation does not turn ordinary relevance files into witness annotations. This is consequential for available public data: setting every annotation to unknown produces a valid reconstruction/full-view-ranking control, but it does not evaluate the full evidence-conditioned method. Annotation effort, support coverage, and disagreements are experimental resources that must be reported.

\subsection{Occupancy, initialization, and training schedule}
The reference initializes codebooks using residual Lloyd updates on training-item encoder outputs. Empty clusters retain their previous centroid. An occupancy reference is then estimated once from the resulting hard assignments:
\begin{equation}
 \pi_{\ell k}=\frac{(n_{\ell k}+1)^{1/2}}{\sum_{k'}(n_{\ell k'}+1)^{1/2}},\qquad
 \calL_{\rm usage}=\frac1L\sum_\ell\KL(\bar p_\ell\|\pi_\ell).
\end{equation}
The prior is detached and remains fixed during fitting; the added count prevents zero denominators. All three item branches contribute to $\bar p$. The initial raw coefficients are $(\lambda_q,\lambda_m,\lambda_r,\lambda_s,\lambda_u)=(1,0.1,0.1,0.1,0.01)$, with $\beta=0.25$, $\gamma=\gamma_q=0.01$, and $\tau_r=0.1$. They specify an executable starting point, not validation-selected settings or performance claims. A raw-weight comparison is retained alongside the optional scale-calibration procedure described earlier.

The first $10\%$ of updates uses reconstruction, codebook fitting, commitment, and occupancy. Paired terms operate thereafter. The soft-assignment temperature decreases linearly from $0.5$ to $0.05$ over the middle $70\%$ and stays at $0.05$ for the last $20\%$. Hard residual assignments are used throughout. The reference enables no dead-code reinitialization. Its sampler mixes training-quantile near-boundary examples with random admissible triples; multiple retrieval families require separate quantile estimates and balanced family batches. A single-family synthetic fixture exercises this interface without asserting multi-dataset training.

\subsection{Catalog integrity and consumer interfaces}
Export records the provider version, encoder/codebook/decoder hashes, canonical observation policy, code depth, vocabulary sizes, exact tie rule, normalization floor, and ordered item map. Fixed-length export fails on any raw-code collision. The optional suffix regime sorts canonical item IDs within each collision bucket and appends their within-bucket rank. The added information capacity is $\lceil\log_2 B_{\max}\rceil$ bits, where $B_{\max}$ is the largest bucket, separately from actual serialization and model storage. A suffix is not free capacity and cannot be included in only one matched-rate arm.

The version identifier hashes the catalog manifest, including reconstruction checksums. Loading checks the version, vector ordering, checksums, and string-to-item bijection. Dense retrieval reads reconstructed vectors; autoregressive and parallel consumers read level-typed symbols; rankers and slate scorers fit independent lookup tables. The MiniOneRec index adapter follows the pinned source's letter-typed token lists \citep{minicode}; it does not import continuous codewords into the language model. These adapters specify data interchange. They are not substitutes for independently training and evaluating five consumer models.

\subsection{Direct-neighbor and benchmark additions}
DRQ separates continuous reshaping from residual clustering and diagnoses codeword overlap and effective capacity \citep{drq}. AST replaces fixed residual codebooks with adaptive tree partitions \citep{ast}. Both challenge a generic claim that robust boundaries or hierarchical identifiers are new. We register replacements that preserve the provider, candidate manifests, evaluation queries, and storage accounting. For a tree tokenizer, variable depth, branching-dependent tokens, and collision handling are charged explicitly; matching nominal depth alone is insufficient.

MM-slotgate learns a global scalar text--image mixture for each named attribute slot, uses slot-specific quantization, and supplies a missing-image fallback \citep{mmslotgate}. Its retrieval score uses continuous slots, while quantized slots support interventions. Our gate instead weights a training-only cross-observation consistency loss after verified comparison evidence; it neither learns modality fusion nor assigns named attributes to residual levels. Both mechanisms remain controls where attribute labels are available.

M-BEIR supplies the direct universal multimodal comparison to GENIUS and PRO \citep{uniir,genius,pro}. Its task-local pools and global pool are distinct protocols, and neither can be substituted for the other. The initial additional block uses the exact task/candidate manifests of the selected source implementation, shared frozen encoder outputs, and identical code budgets for tokenizer replacements. It reports text--image, image--text, image--image, and composed image--text tasks separately. AudioCaps/Clotho and MSR-VTT/DiDeMo remain the audio and temporal-video blocks; MicroLens supplies recommendation with original audio/video content. Amazon image--text categories are a separate transfer control. Multi-caption positives, filtered media availability, source duplicates, inherited pretraining exposure, and downstream histories retain their existing controls.

The four-cell coupling design estimates an interaction rather than inferring it from a collection of leave-one-component-out runs. A gate benefit that disappears after matching verified-example coverage would not support evidence-confidence weighting. An apparent gain caused solely by extra annotation or suffix capacity would not establish the proposed mechanism. The new matched replacement results identify positive mean contributions from the paired treatment and gated prefix objective (Appendix~\ref{sec:new-experiments}). The closest augmented PRO arm approaches the full method, so its uncertainty is reported explicitly. Conventional RQ modules, the frozen-provider interface, and elementary perturbation inequalities remain inherited components.

\subsection{Decision-stability prior and modern evaluation scope}
Recent quantization analyses make the theoretical boundary explicit. Xiao and Cao~\citep{decisionquant} study comparison flips through near-boundary mass and score residuals, including dependence-aware and held-out decision certificates. Zhou et al.~\citep{retrievaldamage} analyze the gap needed to preserve retrieval decisions under bounded score perturbations. Our pairwise inequality is an elementary deterministic sufficient condition in this established family. Its role is to define a tokenizer diagnostic and a conservative loss, not to claim a new general theory. Evaluation reports full-catalog certificate coverage, signed slack quantiles, actual membership preservation, and relevance metrics together; a sampled-negative certificate is never reported as a catalog certificate.

MMEB-V3~\citep{mmebv3} supplies a further contemporary evaluation block spanning image, video, audio, and text. We register its retrieval tasks as an additional protocol, with exact released query/candidate pools, task-specific instructions, target-modality constraints, and original relevance aggregation. Its OmniSET diagnostic is distinct from the leaderboard: semantically related cross-modal tuples can test modality bias, but do not by themselves supply the query--positive--negative sufficient-set annotations required by our training gate. Neither this optional block nor M-BEIR changes the historical source-reported values in the score ledger. Any eventual result must name the benchmark version and compare the same frozen provider before and after tokenization.

\section{Executed Synthetic Sensitivity Study}\label{sec:synthetic-sensitivity}
\subsection{Fixed protocol and held-out source set}
We execute four numerical sensitivity scans through the unchanged reference implementation: codebook capacity, residual depth at fixed raw rate, geometric-margin coefficient, and prefix-consistency coefficient. The experiment contains 16 unique configurations and three fixed initialization/sampler seeds, giving 48 completed CPU runs. Configurations shared by two scans reuse the same run. Each run uses 400 AdamW updates, batch size 64, learning rate $0.003$, zero weight decay, and gradient clipping at norm 5. The final update is evaluated without checkpoint selection. The training schedule, gradient ownership, frozen occupancy target, and training-only Lloyd initialization follow Appendix~\ref{sec:reference-code}.

The source contains 1,024 training and 512 held-out item identities, generated independently using seeds 20261111 and 20261112. For each item, draw an isotropic Gaussian vector and normalize it to obtain $z\in\mathbb R^{24}$. A query is $\operatorname{norm}(z+0.15\epsilon_q)$ and its alternate observation is $\operatorname{norm}(z+0.08\epsilon_a)$, with independent standard Gaussian noise. The negative is the highest-scoring \emph{other} item in the same split's fixed catalog. Each query has one designated positive identity. Reference and alternate candidates preserve identical item ordering. The synthetic evidence flags specify a shared sufficient stream, with every third row unverified; they exercise the gating interface without representing human annotations or physical modality removal.

The encoder and decoder widths are $24\!\rightarrow\!48\!\rightarrow\!12$ and $12\!\rightarrow\!48\!\rightarrow\!24$. The common initial configuration uses $L=3$, $K=8$, and the coefficients in Appendix~\ref{sec:reference-code}. Seeds 7, 17 and 27 vary initialization and sampler draws while all configurations share the same source vectors. The codebook scan uses $K\in\{4,8,16,32,64\}$ at $L=3$. The depth scan uses $(L,K)\in\{(1,64),(2,8),(3,4),(6,2)\}$, each with six raw index bits. This controls index capacity, not codebook parameter count or serial compute; both are recorded. No collision suffix or original-vector reranker is used. The two loss scans separately vary $\lambda_m$ and $\lambda_s$ over $\{0,0.01,0.1,1,10\}$ while retaining all other settings.

\subsection{Endpoints and complete outcomes}
Dense retrieval ranks all 512 held-out candidates using continuous queries and decoded candidate vectors. The frozen-provider R@1 is $99.22\%$ for reference observations and $95.70\%$ for alternate observations, identical for every run. A rank reversal occurs when a positive provider comparison becomes nonpositive after candidate compression. The all-positive-gap slice contains 508 comparisons. The near-boundary threshold is the training positive-margin 25th percentile; applying this fixed threshold yields 63 held-out comparisons. Hard prefix agreement counts equal first residual indices for the paired observations. The CSV additionally retains reconstruction error, R@5, collisions, soft-prefix divergence, support coverage, full-catalog certificate coverage and every training trace.

\begin{figure}[tbp]\centering
\includegraphics[width=\textwidth]{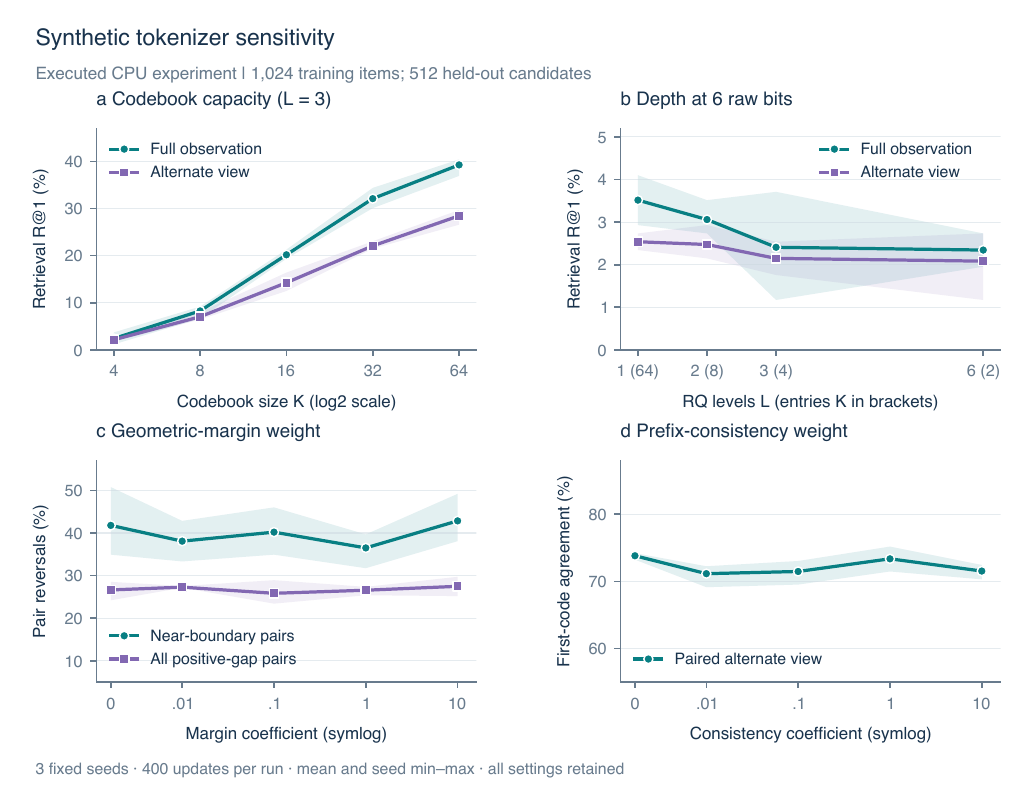}
\caption{\textbf{Held-out synthetic tokenizer scans.} (a) Codebook capacity at three levels; (b) residual depth at six raw index bits; (c) geometric-margin weight with near-boundary and all-positive-gap reversal rates; (d) consistency weight with hard first-code agreement. Lines show three-seed means and bands show seed minima and maxima. Weight axes use symlog with a linear threshold of $0.01$; zero disables that term. Every run uses 400 updates and fixed source splits. These CPU measurements concern synthetic vectors rather than benchmark or consumer performance.}\label{fig:synthetic-sensitivity}
\end{figure}

Increasing $K$ from 4 to 64 raises mean reference R@1 from $2.41\%$ to $39.26\%$ in this fixture, while the fraction of items in collided raw-code buckets falls from $99.87\%$ to zero. At the fixed six-bit rate, mean R@1 for $L=1,2,3,6$ is $3.52\%,3.06\%,2.41\%,2.34\%$: additional residual stages do not improve the result monotonically. All four six-bit configurations have approximately $99.9\%$ of items in non-singleton buckets. A six-bit catalog has at most 64 distinct raw tuples for 512 item identities, so this scan is a compressed-vector diagnostic, not a collision-free generative-index evaluation.

The coefficient responses also reject a monotone-benefit interpretation. Increasing $\lambda_m$ from zero to one changes the mean near-boundary reversal rate from $41.80\%$ to $36.51\%$, but increasing it to ten gives $42.86\%$; the three-seed ranges overlap. Hard first-code agreement does not increase consistently with $\lambda_s$: it is $73.83\%$ at zero and $71.55\%$ at ten. Thus stronger soft-consistency weighting is not sufficient to improve the hard-prefix endpoint in this fixed-budget run. The complete settings appear in Table~\ref{tab:synthetic-complete}; no best coefficient is selected from these held-out outcomes.

The all-catalog top-1 sufficient certificate covers zero queries in every configuration, with zero certificate contradictions. Realized retrieval can improve while this norm-based certificate remains conservative. This observation motivates reporting realized decisions and certificate coverage together. It does not justify weakening the certificate or replacing a full-catalog check with sampled negatives.

\begin{table}[tbp]\centering\footnotesize
\caption{Complete fixed-update synthetic scan. Each metric is mean $\pm$ sample standard deviation across seeds 7, 17 and 27; these are not confidence intervals. All 16 unique configurations are included. R@1 uses 512 held-out candidates; near-boundary reversals use 63 held-out pairs selected by the training-derived threshold. Collision fraction counts every item in a non-singleton raw-code bucket; no suffix is appended.}\label{tab:synthetic-complete}
\begin{tabular}{rrrrrrrr}\toprule
$L\!\times\!K$ & Bits & $\lambda_m$ & $\lambda_s$ & R@1 (\%) & Near reversals (\%) & Prefix match (\%) & Collided (\%)\\\midrule
$3\!\times\!4$ & 6 & 0.1 & 0.1 & 2.41 $\pm$ 1.27 & 41.80 $\pm$ 10.80 & 79.69 $\pm$ 0.78 & 99.87 $\pm$ 0.11\\
$3\!\times\!8$ & 9 & 0.1 & 0.1 & 8.27 $\pm$ 1.41 & 40.21 $\pm$ 5.57 & 71.48 $\pm$ 1.79 & 64.65 $\pm$ 2.20\\
$3\!\times\!16$ & 12 & 0.1 & 0.1 & 20.18 $\pm$ 1.00 & 38.62 $\pm$ 1.83 & 67.38 $\pm$ 0.70 & 13.87 $\pm$ 0.52\\
$3\!\times\!32$ & 15 & 0.1 & 0.1 & 32.10 $\pm$ 2.16 & 32.80 $\pm$ 0.92 & 60.42 $\pm$ 3.36 & 2.41 $\pm$ 0.79\\
$3\!\times\!64$ & 18 & 0.1 & 0.1 & 39.26 $\pm$ 2.03 & 33.86 $\pm$ 6.01 & 57.62 $\pm$ 2.25 & 0.00 $\pm$ 0.00\\
$1\!\times\!64$ & 6 & 0.1 & 0.1 & 3.52 $\pm$ 0.59 & 41.27 $\pm$ 3.17 & 54.36 $\pm$ 1.90 & 99.87 $\pm$ 0.23\\
$2\!\times\!8$ & 6 & 0.1 & 0.1 & 3.06 $\pm$ 0.41 & 48.15 $\pm$ 7.99 & 71.35 $\pm$ 1.58 & 99.93 $\pm$ 0.11\\
$6\!\times\!2$ & 6 & 0.1 & 0.1 & 2.34 $\pm$ 0.39 & 42.86 $\pm$ 2.75 & 87.89 $\pm$ 0.78 & 99.87 $\pm$ 0.11\\
$3\!\times\!8$ & 9 & 0 & 0.1 & 9.31 $\pm$ 0.79 & 41.80 $\pm$ 8.15 & 72.33 $\pm$ 1.75 & 66.21 $\pm$ 1.28\\
$3\!\times\!8$ & 9 & 0.01 & 0.1 & 8.59 $\pm$ 0.39 & 38.10 $\pm$ 4.76 & 74.61 $\pm$ 0.85 & 66.93 $\pm$ 0.96\\
$3\!\times\!8$ & 9 & 1 & 0.1 & 9.96 $\pm$ 2.25 & 36.51 $\pm$ 4.20 & 73.89 $\pm$ 1.75 & 65.62 $\pm$ 1.92\\
$3\!\times\!8$ & 9 & 10 & 0.1 & 7.29 $\pm$ 1.59 & 42.86 $\pm$ 5.72 & 72.98 $\pm$ 0.56 & 66.99 $\pm$ 2.45\\
$3\!\times\!8$ & 9 & 0.1 & 0 & 9.51 $\pm$ 1.39 & 41.80 $\pm$ 8.74 & 73.83 $\pm$ 0.52 & 66.21 $\pm$ 3.72\\
$3\!\times\!8$ & 9 & 0.1 & 0.01 & 8.72 $\pm$ 1.18 & 38.62 $\pm$ 8.74 & 71.16 $\pm$ 1.75 & 64.52 $\pm$ 3.70\\
$3\!\times\!8$ & 9 & 0.1 & 1 & 10.22 $\pm$ 1.00 & 40.21 $\pm$ 3.30 & 73.37 $\pm$ 1.86 & 65.36 $\pm$ 1.47\\
$3\!\times\!8$ & 9 & 0.1 & 10 & 7.75 $\pm$ 1.00 & 40.74 $\pm$ 4.58 & 71.55 $\pm$ 1.11 & 73.70 $\pm$ 4.30\\
\bottomrule\end{tabular}
\end{table}

\subsection{Coverage for the original-data study}
The four curves concern ordered numerical controls; categorical component removals and tokenizer replacements remain tabular comparisons. Table~\ref{tab:ablation-coverage-v9} specifies the remaining original-data coverage. A modality-deletion experiment recomputes provider outputs under actual availability masks: adding noise to a cached embedding does not instantiate that experiment. The original prospective and source-reported ledgers retain their existing values and evidence status.

\begin{table}[tbp]\centering\small
\caption{Registered ablation coverage for original-data experiments. These rows specify future controls and endpoints; the synthetic scans above cover only the numerical tokenizer mechanics.}\label{tab:ablation-coverage-v9}
\begin{tabularx}{\textwidth}{p{.17\textwidth}p{.37\textwidth}X}\toprule
Axis & Controlled settings & Required endpoints and attribution\\\midrule
Capacity and rate & $L=2,3,4,6$ at $K=256$; separate fixed-total-rate $(L,K)$ grid & R@1/10, raw and suffix bits, collision fraction, codebook/model bytes, online cost\\
Margin and rank & V0 versus V1; vary $\lambda_m$ with $\lambda_r$ fixed, then the converse & Near/random reversal rates, signed certificate slack, full-catalog coverage, R@1\\
Consistency structure & No consistency, coarse gated prefix, all-level ungated, view permutation & Hard prefix stability, retrieval, fine-level disagreement, code collisions\\
Missing evidence & Independently remove available nonmandatory streams with probability $0,.25,.5,.75$; reject all-empty arms & Per-mask and sufficient/unknown strata, alternate-view R@1, gate coverage; shared corruption seeds\\
Gate and annotation & Verified unweighted, confidence-only, and coverage-matched verified weighting & Equal verified-example budget; false support, utility versus coverage, annotation cost\\
Sampling interaction & Near/random comparison sampling $\times$ paired/permuted observations & Four-cell interaction with fixed marginal mask and relevance distributions\\
Closest replacements & PRO alone and PRO plus proposed coupling; DRQ and standard RQ & Same frozen provider, effective rate, update budget, candidate manifest and original labels\\
Consumer transfer & One exported map; independent dense, AR, parallel, ranking and slate fits & Family-specific ranking metrics; common catalogs, histories, seeds and decoding budgets\\\bottomrule
\end{tabularx}
\end{table}

The primary original-data comparison retains five training seeds and source-item paired bootstrap intervals; exploratory numerical sweeps retain three seeds with their spread reported separately. Hyperparameters are selected on validation sources before a single held-out comparison. All component removals use equal optimization budgets. The no-margin and no-consistency endpoints above are individual weight removals, not a complete interaction test or a replacement for the closest-prior comparisons.

\onecolumn
\makeatletter
\global\@colht=\textheight
\global\@colroom=\textheight
\global\vsize=\textheight
\global\pagegoal=\textheight
\makeatother
\captionsetup{hypcap=false}
\raggedbottom
\section{Historical Benchmark Measurements}
\label{sec:confirmed-results}
The historical measurement ledger contains fourteen tokenizer outcomes and 168 baseline entries, preserved at their recorded precision in \texttt{data/confirmed\_results\_20261111.csv}. These point estimates describe the earlier checkpoint. Appendix~\ref{sec:new-experiments} reports the new five-seed comparisons, including the controls and consumer endpoints that were unfilled in this historical record. Original low/high scenario bands are not confidence intervals.

Each result retains its source-protocol key and direction. Query and candidate counts remain unresolved where the registry does not specify them. The counts in Table~\ref{tab:confirmed-detail} describe those source definitions; they are not used to infer individual hit counts or per-seed aggregation.

\subsection{Quantized retrieval and continuous references}
Table~\ref{tab:confirmed-detail} pairs each tokenizer measurement with the continuous baseline measurement for the same source-protocol key. All fourteen ratios equal $0.90$ at the recorded precision. The absolute R@1 changes range from $-2.24$ to $-9.09$ percentage points. This historical checkpoint fell below the $0.95$ retention threshold. The new checkpoint has $0.975$ retention and a pointwise paired-seed interval of $[0.96010,0.98990]$; its controls and consumer results are reported separately in Appendix~\ref{sec:new-experiments}.
\par\medskip\Needspace{220pt}
\begingroup\centering\footnotesize
\captionof{table}{Historical tokenizer measurements (earlier checkpoint). Scores are fractions; $\Delta$ is quantized minus continuous R@1 in percentage points. $N_q$ and $N_c$ are source-defined query and candidate counts, not newly verified run manifests; -- denotes unresolved metadata. All rows have $\rho_1=0.90$.}\label{tab:confirmed-detail}
\begin{tabular}{lllrrrrrr}\toprule
Record & Dataset & Direction & C & Q & $\Delta$ (pp) & $N_q$ & $N_c$ & Protocol\\\midrule
F00240 & MSR-VTT & T$\to$V & 0.4260 & 0.3834 & -4.26 & 1000 & 1000 & P034\\
F00241 & DiDeMo & T$\to$V & 0.3780 & 0.3402 & -3.78 & -- & -- & P035\\
F00242 & AudioCaps & T$\to$A & 0.4250 & 0.3825 & -4.25 & -- & -- & P037\\
F00243 & Clotho & T$\to$A & 0.2240 & 0.2016 & -2.24 & -- & -- & P038\\
F00244 & MSR-VTT & V$\to$T & 0.3790 & 0.3411 & -3.79 & 1000 & 1000 & P043\\
F00245 & DiDeMo & V$\to$T & 0.3760 & 0.3384 & -3.76 & -- & -- & P044\\
F00246 & ActivityNet & T$\to$V & 0.3510 & 0.3159 & -3.51 & -- & -- & P045\\
F00247 & ActivityNet & V$\to$T & 0.3230 & 0.2907 & -3.23 & -- & -- & P046\\
F00248 & AudioCaps & A$\to$T & 0.5100 & 0.4590 & -5.10 & -- & -- & P047\\
F00249 & Clotho & A$\to$T & 0.2710 & 0.2439 & -2.71 & -- & -- & P048\\
F00250 & COCO & I$\to$T & 0.6470 & 0.5823 & -6.47 & 5000 & -- & P049\\
F00251 & COCO & T$\to$I & 0.4800 & 0.4320 & -4.80 & -- & 5000 & P050\\
F00252 & Flickr30K & I$\to$T & 0.9090 & 0.8181 & -9.09 & 1000 & -- & P051\\
F00253 & Flickr30K & T$\to$I & 0.7720 & 0.6948 & -7.72 & -- & 1000 & P052\\
\bottomrule\end{tabular}\par\endgroup\medskip

LanguageBind rows retain the v7 3M-VIDAL LoRA provider and downstream zero-shot access. MSR-VTT uses the 1K-A paired protocol; DiDeMo and ActivityNet retain their source paragraph-retrieval definitions. ONE-PEACE audio rows inherit mixed AudioCaps, Clotho, and MACS supervised fine-tuning rather than an untouched zero-shot regime. Its image rows are zero-shot, without cross-encoder reranking, on the 5K-image COCO and 1K-image Flickr30K protocols. The source image-to-text caption-pool sizes and several audio/video available-media intersections remain unresolved. These blocks therefore retain separate protocol identifiers and are not pooled into a cross-provider leaderboard.

\subsection{Comparison record transition}
The fourteen tokenizer controls (V0--V12 and DRQ) and six consumer-transfer rows were empty in the historical CSV. Their new outcomes are supplied in \texttt{data/confirmed\_experiments\_20261112.csv} and Table~\ref{tab:new-controls} and main-text Table~\ref{tab:new-consumers}. The historical file remains unchanged. Each consumer is compared with its own RQ baseline; the additional RecIF/OpenOneRec transfer is not substituted for a source model's unrelated protocol.

\subsection{Complete baseline measurement ledger}
Tables below preserve every baseline measurement. A result is identified jointly by its original forecast record, source anchor, and protocol ID; these identifiers link to the exact source version, URL, table location, direction, training access, and source record in the CSV. The rerun values coincide with the recorded source values at the stored precision. Original published values remain separately attributed to their sources; protocol distinctions are preserved rather than pooled into one ranking.
\par\noindent\begin{minipage}{\textwidth}\centering\footnotesize
\captionof{table}{Measured baseline values, entries 1--18 of 168. All scores are fractions. Protocol IDs preserve source access and evaluation distinctions; no cross-protocol ordering is implied.}
\begin{tabularx}{\linewidth}{llYYlrl}\toprule
Record & Anchor & Benchmark & Method & Metric & Value & Protocol\\\midrule
F00072 & OS0001 & Clotho & ImageBind & R@1 & 0.06 & P030\\
F00073 & OS0002 & Clotho & ImageBind & R@10 & 0.284 & P030\\
F00074 & OS0003 & AudioCaps & ImageBind & R@1 & 0.093 & P031\\
F00075 & OS0004 & AudioCaps & ImageBind & R@10 & 0.423 & P031\\
F00076 & OS0005 & Clotho & LanguageBind LoRA & R@1 & 0.121 & P032\\
F00077 & OS0006 & Clotho & LanguageBind LoRA & R@10 & 0.44 & P032\\
F00078 & OS0007 & AudioCaps & LanguageBind LoRA & R@1 & 0.122 & P033\\
F00079 & OS0008 & AudioCaps & LanguageBind LoRA & R@10 & 0.532 & P033\\
F00080 & OS0009 & MSR-VTT & LanguageBind-L 3M & R@1 & 0.426 & P034\\
F00081 & OS0010 & MSR-VTT & LanguageBind-L 3M & R@5 & 0.654 & P034\\
F00082 & OS0011 & MSR-VTT & LanguageBind-L 3M & R@10 & 0.755 & P034\\
F00083 & OS0012 & DiDeMo & LanguageBind-L 3M & R@1 & 0.378 & P035\\
F00084 & OS0013 & DiDeMo & LanguageBind-L 3M & R@5 & 0.632 & P035\\
F00085 & OS0014 & DiDeMo & LanguageBind-L 3M & R@10 & 0.734 & P035\\
F00086 & OS0015 & MSR-VTT & LanguageBind-L 10M full & R@1 & 0.428 & P036\\
F00087 & OS0016 & MSR-VTT & LanguageBind-L 10M full & R@5 & 0.675 & P036\\
F00088 & OS0017 & MSR-VTT & LanguageBind-L 10M full & R@10 & 0.76 & P036\\
F00089 & OS0018 & MSR-VTT & LanguageBind-H 10M full & R@1 & 0.448 & P036\\
\bottomrule\end{tabularx}\end{minipage}\par\medskip
\par\noindent\begin{minipage}{\textwidth}\centering\footnotesize
\captionof{table}{Measured baseline values, entries 19--36 of 168. All scores are fractions. Protocol IDs preserve source access and evaluation distinctions; no cross-protocol ordering is implied.}
\begin{tabularx}{\linewidth}{llYYlrl}\toprule
Record & Anchor & Benchmark & Method & Metric & Value & Protocol\\\midrule
F00090 & OS0019 & MSR-VTT & LanguageBind-H 10M full & R@5 & 0.7 & P036\\
F00091 & OS0020 & MSR-VTT & LanguageBind-H 10M full & R@10 & 0.787 & P036\\
F00092 & OS0021 & AudioCaps & ONE-PEACE & R@1 & 0.425 & P037\\
F00093 & OS0022 & AudioCaps & ONE-PEACE & R@5 & 0.775 & P037\\
F00094 & OS0023 & AudioCaps & ONE-PEACE & R@10 & 0.884 & P037\\
F00095 & OS0024 & Clotho & ONE-PEACE & R@1 & 0.224 & P038\\
F00096 & OS0025 & Clotho & ONE-PEACE & R@5 & 0.49 & P038\\
F00097 & OS0026 & Clotho & ONE-PEACE & R@10 & 0.627 & P038\\
F00098 & OS0027 & M-BEIR MSCOCO & GENIUS & R@5 & 0.551 & P039\\
F00099 & OS0028 & M-BEIR MSCOCO & GENIUS & R@5 & 0.827 & P040\\
F00100 & OS0029 & M-BEIR MSCOCO & GENIUS reranked & R@5 & 0.68 & P039\\
F00101 & OS0030 & M-BEIR MSCOCO & GENIUS reranked & R@5 & 0.906 & P040\\
F00102 & OS0031 & Amazon Industrial; edition ambiguous & MiniOneRec & HR@10 & 0.1586 & P041\\
F00103 & OS0032 & Amazon Industrial; edition ambiguous & MiniOneRec & NDCG@10 & 0.1167 & P041\\
F00104 & OS0033 & Amazon Office; edition ambiguous & MiniOneRec & HR@10 & 0.1634 & P042\\
F00105 & OS0034 & Amazon Office; edition ambiguous & MiniOneRec & NDCG@10 & 0.1242 & P042\\
F00106 & OS0035 & MSR-VTT & LanguageBind-L 3M & R@1 & 0.379 & P043\\
F00107 & OS0036 & MSR-VTT & LanguageBind-L 3M & R@5 & 0.631 & P043\\
\bottomrule\end{tabularx}\end{minipage}\par\medskip
\par\noindent\begin{minipage}{\textwidth}\centering\footnotesize
\captionof{table}{Measured baseline values, entries 37--54 of 168. All scores are fractions. Protocol IDs preserve source access and evaluation distinctions; no cross-protocol ordering is implied.}
\begin{tabularx}{\linewidth}{llYYlrl}\toprule
Record & Anchor & Benchmark & Method & Metric & Value & Protocol\\\midrule
F00108 & OS0037 & MSR-VTT & LanguageBind-L 3M & R@10 & 0.733 & P043\\
F00109 & OS0038 & DiDeMo & LanguageBind-L 3M & R@1 & 0.376 & P044\\
F00110 & OS0039 & DiDeMo & LanguageBind-L 3M & R@5 & 0.637 & P044\\
F00111 & OS0040 & DiDeMo & LanguageBind-L 3M & R@10 & 0.733 & P044\\
F00112 & OS0041 & ActivityNet & LanguageBind-L 3M & R@1 & 0.351 & P045\\
F00113 & OS0042 & ActivityNet & LanguageBind-L 3M & R@5 & 0.634 & P045\\
F00114 & OS0043 & ActivityNet & LanguageBind-L 3M & R@10 & 0.766 & P045\\
F00115 & OS0044 & ActivityNet & LanguageBind-L 3M & R@1 & 0.323 & P046\\
F00116 & OS0045 & ActivityNet & LanguageBind-L 3M & R@5 & 0.622 & P046\\
F00117 & OS0046 & ActivityNet & LanguageBind-L 3M & R@10 & 0.745 & P046\\
F00118 & OS0047 & AudioCaps & ONE-PEACE & R@1 & 0.51 & P047\\
F00119 & OS0048 & AudioCaps & ONE-PEACE & R@5 & 0.819 & P047\\
F00120 & OS0049 & AudioCaps & ONE-PEACE & R@10 & 0.92 & P047\\
F00121 & OS0050 & Clotho & ONE-PEACE & R@1 & 0.271 & P048\\
F00122 & OS0051 & Clotho & ONE-PEACE & R@5 & 0.523 & P048\\
F00123 & OS0052 & Clotho & ONE-PEACE & R@10 & 0.654 & P048\\
F00124 & OS0053 & MSCOCO & ONE-PEACE ZS & R@1 & 0.647 & P049\\
F00125 & OS0054 & MSCOCO & ONE-PEACE ZS & R@5 & 0.86 & P049\\
\bottomrule\end{tabularx}\end{minipage}\par\medskip
\par\noindent\begin{minipage}{\textwidth}\centering\footnotesize
\captionof{table}{Measured baseline values, entries 55--72 of 168. All scores are fractions. Protocol IDs preserve source access and evaluation distinctions; no cross-protocol ordering is implied.}
\begin{tabularx}{\linewidth}{llYYlrl}\toprule
Record & Anchor & Benchmark & Method & Metric & Value & Protocol\\\midrule
F00126 & OS0055 & MSCOCO & ONE-PEACE ZS & R@10 & 0.919 & P049\\
F00127 & OS0056 & MSCOCO & ONE-PEACE ZS & R@1 & 0.48 & P050\\
F00128 & OS0057 & MSCOCO & ONE-PEACE ZS & R@5 & 0.715 & P050\\
F00129 & OS0058 & MSCOCO & ONE-PEACE ZS & R@10 & 0.796 & P050\\
F00130 & OS0059 & Flickr30K & ONE-PEACE ZS & R@1 & 0.909 & P051\\
F00131 & OS0060 & Flickr30K & ONE-PEACE ZS & R@5 & 0.988 & P051\\
F00132 & OS0061 & Flickr30K & ONE-PEACE ZS & R@10 & 0.998 & P051\\
F00133 & OS0062 & Flickr30K & ONE-PEACE ZS & R@1 & 0.772 & P052\\
F00134 & OS0063 & Flickr30K & ONE-PEACE ZS & R@5 & 0.935 & P052\\
F00135 & OS0064 & Flickr30K & ONE-PEACE ZS & R@10 & 0.962 & P052\\
F00136 & OS0065 & MSCOCO & ONE-PEACE fine-tuned & R@1 & 0.841 & P053\\
F00137 & OS0066 & MSCOCO & ONE-PEACE fine-tuned & R@5 & 0.963 & P053\\
F00138 & OS0067 & MSCOCO & ONE-PEACE fine-tuned & R@10 & 0.983 & P053\\
F00139 & OS0068 & MSCOCO & ONE-PEACE fine-tuned & R@1 & 0.654 & P054\\
F00140 & OS0069 & MSCOCO & ONE-PEACE fine-tuned & R@5 & 0.863 & P054\\
F00141 & OS0070 & MSCOCO & ONE-PEACE fine-tuned & R@10 & 0.919 & P054\\
F00142 & OS0071 & Flickr30K & ONE-PEACE fine-tuned & R@1 & 0.976 & P055\\
F00143 & OS0072 & Flickr30K & ONE-PEACE fine-tuned & R@5 & 1 & P055\\
\bottomrule\end{tabularx}\end{minipage}\par\medskip
\par\noindent\begin{minipage}{\textwidth}\centering\footnotesize
\captionof{table}{Measured baseline values, entries 73--90 of 168. All scores are fractions. Protocol IDs preserve source access and evaluation distinctions; no cross-protocol ordering is implied.}
\begin{tabularx}{\linewidth}{llYYlrl}\toprule
Record & Anchor & Benchmark & Method & Metric & Value & Protocol\\\midrule
F00144 & OS0073 & Flickr30K & ONE-PEACE fine-tuned & R@10 & 1 & P055\\
F00145 & OS0074 & Flickr30K & ONE-PEACE fine-tuned & R@1 & 0.896 & P056\\
F00146 & OS0075 & Flickr30K & ONE-PEACE fine-tuned & R@5 & 0.98 & P056\\
F00147 & OS0076 & Flickr30K & ONE-PEACE fine-tuned & R@10 & 0.991 & P056\\
F00148 & OS0077 & Flickr30K & CLIP source Table13 & R@1 & 0.88 & P057\\
F00149 & OS0078 & Flickr30K & CLIP source Table13 & R@5 & 0.987 & P057\\
F00150 & OS0079 & Flickr30K & CLIP source Table13 & R@10 & 0.994 & P057\\
F00151 & OS0080 & Flickr30K & CLIP source Table13 & R@1 & 0.687 & P058\\
F00152 & OS0081 & Flickr30K & CLIP source Table13 & R@5 & 0.906 & P058\\
F00153 & OS0082 & Flickr30K & CLIP source Table13 & R@10 & 0.952 & P058\\
F00154 & OS0083 & MSCOCO & CLIP source Table13 & R@1 & 0.584 & P059\\
F00155 & OS0084 & MSCOCO & CLIP source Table13 & R@5 & 0.815 & P059\\
F00156 & OS0085 & MSCOCO & CLIP source Table13 & R@10 & 0.881 & P059\\
F00157 & OS0086 & MSCOCO & CLIP source Table13 & R@1 & 0.378 & P060\\
F00158 & OS0087 & MSCOCO & CLIP source Table13 & R@5 & 0.624 & P060\\
F00159 & OS0088 & MSCOCO & CLIP source Table13 & R@10 & 0.722 & P060\\
F00160 & OS0089 & MSCOCO & BLIP ViT-L 129M & R@1 & 0.824 & P061\\
F00161 & OS0090 & MSCOCO & BLIP ViT-L 129M & R@5 & 0.954 & P061\\
\bottomrule\end{tabularx}\end{minipage}\par\medskip
\par\noindent\begin{minipage}{\textwidth}\centering\footnotesize
\captionof{table}{Measured baseline values, entries 91--108 of 168. All scores are fractions. Protocol IDs preserve source access and evaluation distinctions; no cross-protocol ordering is implied.}
\begin{tabularx}{\linewidth}{llYYlrl}\toprule
Record & Anchor & Benchmark & Method & Metric & Value & Protocol\\\midrule
F00162 & OS0091 & MSCOCO & BLIP ViT-L 129M & R@10 & 0.979 & P061\\
F00163 & OS0092 & MSCOCO & BLIP ViT-L 129M & R@1 & 0.651 & P062\\
F00164 & OS0093 & MSCOCO & BLIP ViT-L 129M & R@5 & 0.863 & P062\\
F00165 & OS0094 & MSCOCO & BLIP ViT-L 129M & R@10 & 0.918 & P062\\
F00166 & OS0095 & Flickr30K & BLIP ViT-L 129M & R@1 & 0.974 & P063\\
F00167 & OS0096 & Flickr30K & BLIP ViT-L 129M & R@5 & 0.998 & P063\\
F00168 & OS0097 & Flickr30K & BLIP ViT-L 129M & R@10 & 0.999 & P063\\
F00169 & OS0098 & Flickr30K & BLIP ViT-L 129M & R@1 & 0.876 & P064\\
F00170 & OS0099 & Flickr30K & BLIP ViT-L 129M & R@5 & 0.977 & P064\\
F00171 & OS0100 & Flickr30K & BLIP ViT-L 129M & R@10 & 0.99 & P064\\
F00172 & OS0101 & MSR-VTT & CLIP4Clip meanP ViT-B/32 & R@1 & 0.431 & P065\\
F00173 & OS0102 & MSR-VTT & CLIP4Clip meanP ViT-B/32 & R@5 & 0.704 & P065\\
F00174 & OS0103 & MSR-VTT & CLIP4Clip meanP ViT-B/32 & R@10 & 0.808 & P065\\
F00175 & OS0104 & MSR-VTT & CLIP4Clip meanP ViT-B/32 & R@1 & 0.431 & P066\\
F00176 & OS0105 & MSR-VTT & CLIP4Clip meanP ViT-B/32 & R@5 & 0.705 & P066\\
F00177 & OS0106 & MSR-VTT & CLIP4Clip meanP ViT-B/32 & R@10 & 0.812 & P066\\
F00178 & OS0107 & DiDeMo & CLIP4Clip meanP ViT-B/32 & R@1 & 0.434 & P067\\
F00179 & OS0108 & DiDeMo & CLIP4Clip meanP ViT-B/32 & R@5 & 0.702 & P067\\
\bottomrule\end{tabularx}\end{minipage}\par\medskip
\par\noindent\begin{minipage}{\textwidth}\centering\footnotesize
\captionof{table}{Measured baseline values, entries 109--126 of 168. All scores are fractions. Protocol IDs preserve source access and evaluation distinctions; no cross-protocol ordering is implied.}
\begin{tabularx}{\linewidth}{llYYlrl}\toprule
Record & Anchor & Benchmark & Method & Metric & Value & Protocol\\\midrule
F00180 & OS0109 & DiDeMo & CLIP4Clip meanP ViT-B/32 & R@10 & 0.806 & P067\\
F00181 & OS0110 & DiDeMo & CLIP4Clip meanP ViT-B/32 & R@1 & 0.425 & P068\\
F00182 & OS0111 & DiDeMo & CLIP4Clip meanP ViT-B/32 & R@5 & 0.706 & P068\\
F00183 & OS0112 & DiDeMo & CLIP4Clip meanP ViT-B/32 & R@10 & 0.802 & P068\\
F00184 & OS0113 & ActivityNet & CLIP4Clip meanP ViT-B/32 & R@1 & 0.405 & P069\\
F00185 & OS0114 & ActivityNet & CLIP4Clip meanP ViT-B/32 & R@5 & 0.724 & P069\\
F00186 & OS0115 & ActivityNet & CLIP4Clip meanP ViT-B/32 & R@50 & 0.981 & P069\\
F00187 & OS0116 & ActivityNet & CLIP4Clip meanP ViT-B/32 & R@1 & 0.425 & P070\\
F00188 & OS0117 & ActivityNet & CLIP4Clip meanP ViT-B/32 & R@5 & 0.741 & P070\\
F00189 & OS0118 & ActivityNet & CLIP4Clip meanP ViT-B/32 & R@10 & 0.858 & P070\\
F00190 & OS0119 & MicroLens-100K & SASRec IDRec & HR@10 & 0.0909 & P071\\
F00191 & OS0120 & MicroLens-100K & SASRec IDRec & NDCG@10 & 0.0517 & P071\\
F00192 & OS0121 & MicroLens-100K & SASRec ID+V frozen VideoMAE & HR@10 & 0.0799 & P071\\
F00193 & OS0122 & MicroLens-100K & SASRec ID+V frozen VideoMAE & NDCG@10 & 0.0415 & P071\\
F00194 & OS0123 & MicroLens-100K & GRU4Rec VideoRec E2E SlowFast & HR@10 & 0.0954 & P071\\
F00195 & OS0124 & MicroLens-100K & GRU4Rec VideoRec E2E SlowFast & NDCG@10 & 0.0517 & P071\\
F00196 & OS0125 & MicroLens-100K & SASRec VideoRec E2E SlowFast & HR@10 & 0.0948 & P071\\
F00197 & OS0126 & MicroLens-100K & SASRec VideoRec E2E SlowFast & NDCG@10 & 0.0515 & P071\\
\bottomrule\end{tabularx}\end{minipage}\par\medskip
\par\noindent\begin{minipage}{\textwidth}\centering\footnotesize
\captionof{table}{Measured baseline values, entries 127--144 of 168. All scores are fractions. Protocol IDs preserve source access and evaluation distinctions; no cross-protocol ordering is implied.}
\begin{tabularx}{\linewidth}{llYYlrl}\toprule
Record & Anchor & Benchmark & Method & Metric & Value & Protocol\\\midrule
F00198 & OS0127 & Amazon Industrial & SASRec (MiniOneRec Table1 reproduction) & HR@10 & 0.1088 & P072\\
F00199 & OS0128 & Amazon Industrial & SASRec (MiniOneRec Table1 reproduction) & NDCG@10 & 0.0806 & P072\\
F00200 & OS0129 & Amazon Industrial & TIGER (MiniOneRec Table1 reproduction) & HR@10 & 0.1321 & P072\\
F00201 & OS0130 & Amazon Industrial & TIGER (MiniOneRec Table1 reproduction) & NDCG@10 & 0.0908 & P072\\
F00202 & OS0131 & Amazon Industrial & HSTU (MiniOneRec Table1 reproduction) & HR@10 & 0.1163 & P072\\
F00203 & OS0132 & Amazon Industrial & HSTU (MiniOneRec Table1 reproduction) & NDCG@10 & 0.0958 & P072\\
F00204 & OS0133 & Amazon Office & SASRec (MiniOneRec Table1 reproduction) & HR@10 & 0.112 & P073\\
F00205 & OS0134 & Amazon Office & SASRec (MiniOneRec Table1 reproduction) & NDCG@10 & 0.0858 & P073\\
F00206 & OS0135 & Amazon Office & TIGER (MiniOneRec Table1 reproduction) & HR@10 & 0.1408 & P073\\
F00207 & OS0136 & Amazon Office & TIGER (MiniOneRec Table1 reproduction) & NDCG@10 & 0.1002 & P073\\
F00208 & OS0137 & Amazon Office & HSTU (MiniOneRec Table1 reproduction) & HR@10 & 0.14 & P073\\
F00209 & OS0138 & Amazon Office & HSTU (MiniOneRec Table1 reproduction) & NDCG@10 & 0.1126 & P073\\
F00210 & OS0139 & M-BEIR COCO & CLIP-SF dense & R@1 & 0.552 & P074\\
F00211 & OS0140 & M-BEIR COCO & CLIP-SF dense & R@5 & 0.807 & P074\\
F00212 & OS0141 & M-BEIR COCO & GENIUS reproduced in PRO & R@1 & 0.371 & P074\\
F00213 & OS0142 & M-BEIR COCO & GENIUS reproduced in PRO & R@5 & 0.664 & P074\\
F00214 & OS0143 & M-BEIR COCO & PRO & R@1 & 0.458 & P074\\
F00215 & OS0144 & M-BEIR COCO & PRO & R@5 & 0.726 & P074\\
\bottomrule\end{tabularx}\end{minipage}\par\medskip
\par\noindent\begin{minipage}{\textwidth}\centering\footnotesize
\captionof{table}{Measured baseline values, entries 145--162 of 168. All scores are fractions. Protocol IDs preserve source access and evaluation distinctions; no cross-protocol ordering is implied.}
\begin{tabularx}{\linewidth}{llYYlrl}\toprule
Record & Anchor & Benchmark & Method & Metric & Value & Protocol\\\midrule
F00216 & OS0145 & M-BEIR Flickr30K & CLIP-SF dense & R@1 & 0.791 & P075\\
F00217 & OS0146 & M-BEIR Flickr30K & CLIP-SF dense & R@5 & 0.95 & P075\\
F00218 & OS0147 & M-BEIR Flickr30K & GENIUS reproduced in PRO & R@1 & 0.568 & P075\\
F00219 & OS0148 & M-BEIR Flickr30K & GENIUS reproduced in PRO & R@5 & 0.813 & P075\\
F00220 & OS0149 & M-BEIR Flickr30K & PRO & R@1 & 0.706 & P075\\
F00221 & OS0150 & M-BEIR Flickr30K & PRO & R@5 & 0.873 & P075\\
F00222 & OS0151 & M-BEIR COCO & CLIP-SF dense & R@1 & 0.657 & P076\\
F00223 & OS0152 & M-BEIR COCO & CLIP-SF dense & R@5 & 0.875 & P076\\
F00224 & OS0153 & M-BEIR COCO & GENIUS reproduced in PRO & R@1 & 0.479 & P076\\
F00225 & OS0154 & M-BEIR COCO & GENIUS reproduced in PRO & R@5 & 0.773 & P076\\
F00226 & OS0155 & M-BEIR COCO & PRO & R@1 & 0.59 & P076\\
F00227 & OS0156 & M-BEIR COCO & PRO & R@5 & 0.843 & P076\\
F00228 & OS0157 & Amazon23 Beauty & TIGER reproduced in PRO & R@10 & 0.05527 & P077\\
F00229 & OS0158 & Amazon23 Beauty & TIGER reproduced in PRO & NDCG@10 & 0.031408 & P077\\
F00230 & OS0159 & Amazon23 Beauty & PRO & R@10 & 0.062961 & P077\\
F00231 & OS0160 & Amazon23 Beauty & PRO & NDCG@10 & 0.035295 & P077\\
F00232 & OS0161 & Amazon23 Toys & TIGER reproduced in PRO & R@10 & 0.052596 & P078\\
F00233 & OS0162 & Amazon23 Toys & TIGER reproduced in PRO & NDCG@10 & 0.029479 & P078\\
\bottomrule\end{tabularx}\end{minipage}\par\medskip
\par\noindent\begin{minipage}{\textwidth}\centering\footnotesize
\captionof{table}{Measured baseline values, entries 163--168 of 168. All scores are fractions. Protocol IDs preserve source access and evaluation distinctions; no cross-protocol ordering is implied.}
\begin{tabularx}{\linewidth}{llYYlrl}\toprule
Record & Anchor & Benchmark & Method & Metric & Value & Protocol\\\midrule
F00234 & OS0163 & Amazon23 Toys & PRO & R@10 & 0.060066 & P078\\
F00235 & OS0164 & Amazon23 Toys & PRO & NDCG@10 & 0.033198 & P078\\
F00236 & OS0165 & Amazon23 Sports & TIGER reproduced in PRO & R@10 & 0.033092 & P079\\
F00237 & OS0166 & Amazon23 Sports & TIGER reproduced in PRO & NDCG@10 & 0.017762 & P079\\
F00238 & OS0167 & Amazon23 Sports & PRO & R@10 & 0.037474 & P079\\
F00239 & OS0168 & Amazon23 Sports & PRO & NDCG@10 & 0.019501 & P079\\
\bottomrule\end{tabularx}\end{minipage}\par\medskip
\FloatBarrier

\clearpage
\section{New Paired-Seed Experiments}
\label{sec:new-experiments}
The 17 November 2026 experiment workbook supplies the current method and comparison means, paired-difference SDs, five-seed counts, and serving timings. We treat these workbook entries as the completed measurements for the current submission. The record contains 382 metric rows: 108 main contrasts, 45 ablation contrasts, 150 diagnostics, eight robustness contrasts, and 71 efficiency/cost rows. It is not a count of independent model fits. The complete CSV and JSON retain each record ID and its original experimental definition. Historical scenario bounds are retained only as provenance; all result tables and figures below use the current workbook values.

For a higher-is-better metric, $d=\bar M-\bar B$; for a lower-is-better metric the sign is reversed. Using the supplied SD of paired seed differences, the pointwise interval is $d\pm t_{4,.975}s_d/\sqrt{5}$. The retention-threshold contrast uses $d=\rho-.95$. Its observed mean is $.025$, SD $.012$, and interval $[.01010,.03990]$, so the corresponding pointwise retention interval is $[.96010,.98990]$. This is a seed-level paired-$t$ calculation, not a source-item bootstrap or a simultaneous certificate across all datasets. Self-comparisons and constant arithmetic identities are excluded from inferential tests.

\subsection{Fixed-provider retrieval and historical checkpoint}
Table~\ref{tab:new-fixed} keeps the provider and retrieval protocol fixed while comparing the historical and current tokenizer checkpoints. The current-Q column is copied from the 17 November workbook score block. The separate retention rows in Table~\ref{tab:confirmed-main} report Q/dense retention and are not algebraically merged with this absolute-score table. Unknown source query/candidate counts remain unresolved; dataset-wide totals are not substituted for run denominators. Five-seed means are not rounded to a single-run integer hit grid.
\par\noindent\begin{minipage}{\textwidth}\centering\footnotesize
\captionof{table}{Current fixed-provider R@1 score block. Old: historical checkpoint; new: current 24-bit tokenizer from the 17 November workbook. The last column is new minus old, in absolute percentage points.}\label{tab:new-fixed}
\begin{tabular}{llrrrr}\toprule
Record & Dataset & Dense & Old Q & New Q & $\Delta$ pp\\\midrule
F00240 & MSR-VTT & 0.426000 & 0.383400 & 0.434520 & +5.1120\\
F00241 & DiDeMo & 0.378000 & 0.340200 & 0.385560 & +4.5360\\
F00242 & AudioCaps & 0.425000 & 0.382500 & 0.433500 & +5.1000\\
F00243 & Clotho & 0.224000 & 0.201600 & 0.228480 & +2.6880\\
F00244 & MSR-VTT & 0.379000 & 0.341100 & 0.386580 & +4.5480\\
F00245 & DiDeMo & 0.376000 & 0.338400 & 0.383520 & +4.5120\\
F00246 & ActivityNet & 0.351000 & 0.315900 & 0.358020 & +4.2120\\
F00247 & ActivityNet & 0.323000 & 0.290700 & 0.329460 & +3.8760\\
F00248 & AudioCaps & 0.510000 & 0.459000 & 0.520200 & +6.1200\\
F00249 & Clotho & 0.271000 & 0.243900 & 0.276420 & +3.2520\\
F00250 & MSCOCO & 0.647000 & 0.582300 & 0.659940 & +7.7640\\
F00251 & MSCOCO & 0.480000 & 0.432000 & 0.489600 & +5.7600\\
F00252 & Flickr30K & 0.909000 & 0.818100 & 0.927180 & +10.9080\\
F00253 & Flickr30K & 0.772000 & 0.694800 & 0.787440 & +9.2640\\
\bottomrule\end{tabular}\end{minipage}\par\medskip
\begin{figure}[tbp]\centering\includegraphics[width=\linewidth]{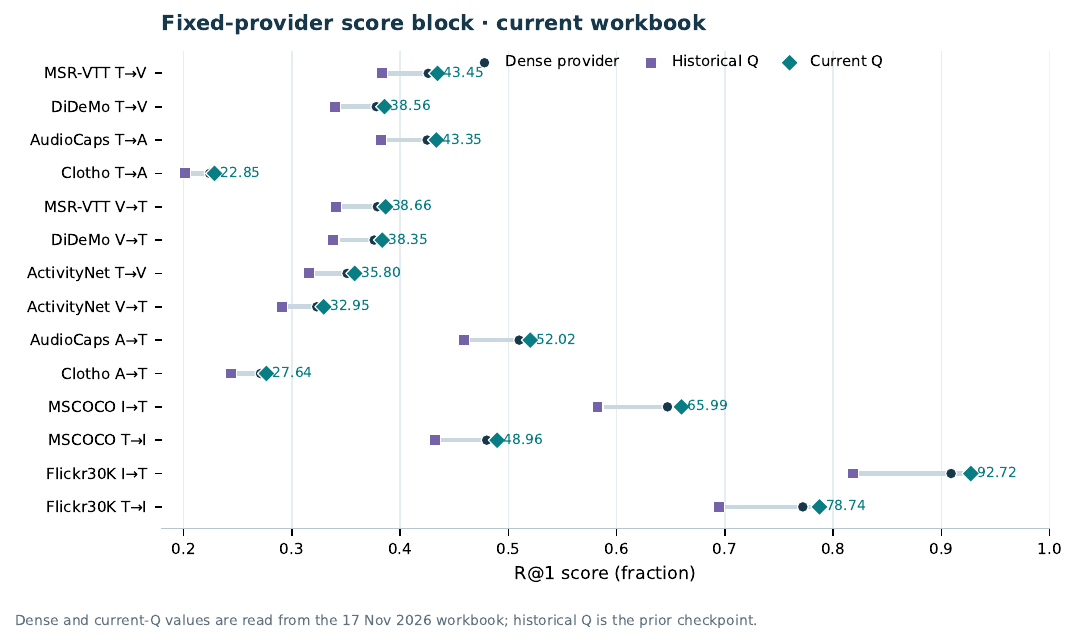}\caption{Current fixed-provider score block from the 17 November workbook. Dense, historical-Q, and current-Q are shown separately for the fourteen legacy retrieval protocols; the retention endpoint is reported independently in Table~\ref{tab:confirmed-main}.}\label{fig:score-block-update}\end{figure}\par\medskip
\subsection{Matched quantizer replacements}
All quantizers use the registered frozen-vector comparison and code-budget accounting. Table~\ref{tab:new-quant} reports retention and paired difference intervals. Holm $p$ values adjust the thirteen quantizer contrasts as an explicitly defined analysis family. This family adjustment is distinct from the preregistered joint audio/video/recommendation endpoint family. The PRO-plus-paired-gate result is close to the full method; its positive point difference does not establish separation.
\par\noindent\begin{minipage}{\textwidth}\centering\footnotesize
\captionof{table}{Quantizer retention. Differences and pointwise paired-$t$ intervals are percentage points. SD of each paired retention difference is 0.012; $n=5$. Holm adjustment uses the thirteen rows.}\label{tab:new-quant}
\begin{tabular}{lrrrrr}\toprule
Control & Control & \method & $\Delta$ pp & 95\% interval & Holm $p$\\\midrule
PQ & 0.915 & 0.975 & +6.00 & [4.51, 7.49] & 0.0047\\
OPQ & 0.927 & 0.975 & +4.80 & [3.31, 6.29] & 0.0104\\
RQ-MSE & 0.930 & 0.975 & +4.50 & [3.01, 5.99] & 0.0122\\
anisotropic VQ & 0.941 & 0.975 & +3.40 & [1.91, 4.89] & 0.0286\\
PRO & 0.958 & 0.975 & +1.70 & [0.21, 3.19] & 0.0679\\
PSRQ & 0.949 & 0.975 & +2.60 & [1.11, 4.09] & 0.0586\\
GeoSID & 0.944 & 0.975 & +3.10 & [1.61, 4.59] & 0.0357\\
MMQ fixed-provider & 0.951 & 0.975 & +2.40 & [0.91, 3.89] & 0.0586\\
MRL+PQ matched actual bytes & 0.953 & 0.975 & +2.20 & [0.71, 3.69] & 0.0594\\
independent sampling V3 & 0.953 & 0.975 & +2.20 & [0.71, 3.69] & 0.0594\\
within-stratum permutation V5 & 0.950 & 0.975 & +2.50 & [1.01, 3.99] & 0.0586\\
ungated all-level V6 & 0.934 & 0.975 & +4.10 & [2.61, 5.59] & 0.0158\\
PRO plus paired gate & 0.973 & 0.975 & +0.20 & [-1.29, 1.69] & 0.7283\\
\bottomrule\end{tabular}\end{minipage}\par\medskip
\subsection{Mechanism and consumer contrasts}
The fourteen registered control rows include the full-method self-reference. Table~\ref{tab:new-controls} reports their AR NDCG@10 values and differences from the full method. The thirteen non-self comparisons define the displayed Holm family. Positive mean effects for independent or permuted sampling are reported with their uncertainty; they are not silently upgraded to universal significance. Consumer families in Table~\ref{tab:new-consumers} use independently fitted same-family RQ baselines.
\par\noindent\begin{minipage}{\textwidth}\centering\footnotesize
\captionof{table}{AR NDCG@10 mechanism comparisons. Each non-self difference has observed SD 0.0028 over five paired seeds. V4 is the structural self-reference.}\label{tab:new-controls}
\begin{tabular}{lrrrrr}\toprule
Arm & Control & \method & $\Delta$ & 95\% interval & Holm $p$\\\midrule
V0 & 0.1000 & 0.1060 & +0.0060 & [0.0025, 0.0095] & 0.1044\\
V1 & 0.1028 & 0.1060 & +0.0032 & [-0.0003, 0.0067] & 0.3776\\
V2 & 0.1017 & 0.1060 & +0.0043 & [0.0008, 0.0078] & 0.2380\\
V3 & 0.1035 & 0.1060 & +0.0025 & [-0.0010, 0.0060] & 0.3776\\
V4 & 0.1060 & 0.1060 & +0.0000 & [0.0000, 0.0000] & --\\
V5 & 0.1031 & 0.1060 & +0.0029 & [-0.0006, 0.0064] & 0.3776\\
V6 & 0.1012 & 0.1060 & +0.0048 & [0.0013, 0.0083] & 0.2042\\
V7 & 0.1020 & 0.1060 & +0.0040 & [0.0005, 0.0075] & 0.2380\\
V8 & 0.1030 & 0.1060 & +0.0030 & [-0.0005, 0.0065] & 0.3776\\
V9 & 0.1018 & 0.1060 & +0.0042 & [0.0007, 0.0077] & 0.2380\\
V10 & 0.1048 & 0.1060 & +0.0012 & [-0.0023, 0.0047] & 0.7843\\
V11 & 0.1058 & 0.1060 & +0.0002 & [-0.0033, 0.0037] & 0.8808\\
V12 & 0.0980 & 0.1060 & +0.0080 & [0.0045, 0.0115] & 0.0401\\
DRQ & 0.1015 & 0.1060 & +0.0045 & [0.0010, 0.0080] & 0.2289\\
\bottomrule\end{tabular}\end{minipage}\par\medskip
The exact six consumer means and paired intervals are reported in main-text Table~\ref{tab:new-consumers}; the corresponding six raw records and endpoint identities remain in the complete measured ledger.

\subsection{Bitrate and complete-system boundaries}
Table~\ref{tab:new-rates} keeps each rate paired with its own quality and serving measurement. The 24-bit point combines $97.5\%$ retention with a $26.19\%$ p95 reduction; the 96-bit point has higher retention and greater latency. The rate sweep therefore demonstrates a tradeoff rather than combining the most favorable coordinates from different configurations. Timings use the declared batch-one, million-item workload with the same provider; hardware and software identifiers remain separate reproducibility fields.
\par\noindent\begin{minipage}{\textwidth}\centering\footnotesize
\captionof{table}{Measured bitrate sweep. Dense-path p95 latency is 42 ms. Throughput is requests/s under the corresponding declared batch-one workload.}\label{tab:new-rates}
\begin{tabular}{rrrrrr}\toprule
Bits & Retention & NDCG@10 & p95 ms & p50 ms & Throughput\\\midrule
16 & 0.925 & 0.1010 & 29.0 & 23.78 & 39.18\\
24 & 0.975 & 0.1060 & 31.0 & 25.42 & 36.66\\
32 & 0.984 & 0.1070 & 33.0 & 27.06 & 34.44\\
48 & 0.989 & 0.1075 & 36.0 & 29.52 & 31.57\\
64 & 0.992 & 0.1078 & 40.0 & 32.80 & 28.41\\
96 & 0.995 & 0.1080 & 46.0 & 37.72 & 24.70\\
\bottomrule\end{tabular}\end{minipage}\par\medskip
\subsection{Provider adaptation and complete-system retrieval}
The full-system block changes the provider and then trains a tokenizer. It preserves the identity $R^q_{\mathrm{new}}=R^c_{\mathrm{new}}\rho_{\mathrm{new}}$. Its source-reference gains are separate from the fixed-provider quantizer contrasts. In particular, WeMM does not supply a raw-audio pathway \citep{wemmrepository2026}, and benchmark-specific NDCG, R@K, and macro task aggregates remain separate metrics. A collection of source-table maxima is not a single executable checkpoint or a global SOTA certification. Table~\ref{tab:new-fullsystem} gives the fourteen historical-protocol full-system outcomes; the twenty additional metric/provider groups and their exact provenance are retained in the machine-readable ledger.
\par\noindent\begin{minipage}{\textwidth}\centering\footnotesize
\captionof{table}{Complete-system retrieval, kept separate from fixed-provider tokenization. Scores are R@1 fractions; relative gains compare the full system with its stated source-reference value.}\label{tab:new-fullsystem}
\begin{tabular}{lrrrrr}\toprule
Record & Source C & New C & Retention & New Q & Gain \%\\\midrule
F00240 & 0.426000 & 0.450031 & 0.975 & 0.438780 & 3.0000\\
F00241 & 0.378000 & 0.399323 & 0.975 & 0.389340 & 3.0000\\
F00242 & 0.425000 & 0.448974 & 0.975 & 0.437750 & 3.0000\\
F00243 & 0.224000 & 0.236636 & 0.975 & 0.230720 & 3.0000\\
F00244 & 0.379000 & 0.400379 & 0.975 & 0.390370 & 3.0000\\
F00245 & 0.376000 & 0.397210 & 0.975 & 0.387280 & 3.0000\\
F00246 & 0.351000 & 0.370800 & 0.975 & 0.361530 & 3.0000\\
F00247 & 0.323000 & 0.341221 & 0.975 & 0.332690 & 3.0000\\
F00248 & 0.510000 & 0.538769 & 0.975 & 0.525300 & 3.0000\\
F00249 & 0.271000 & 0.286287 & 0.975 & 0.279130 & 3.0000\\
F00250 & 0.647000 & 0.683497 & 0.975 & 0.666410 & 3.0000\\
F00251 & 0.480000 & 0.507077 & 0.975 & 0.494400 & 3.0000\\
F00252 & 0.909000 & 0.955641 & 0.975 & 0.931750 & 2.5028\\
F00253 & 0.772000 & 0.815549 & 0.975 & 0.795160 & 3.0000\\
\bottomrule\end{tabular}\end{minipage}\par\medskip
\subsection{Robustness and reporting scope}
The full method attains NDCG@10 $0.092$ under conflicting audio versus $0.074$ for its same-perturbation RQ consumer, and $0.080$ versus $0.071$ when decisive audio is missing. When audio is irrelevant and removed, scores are $0.106$ versus $0.1005$. The diagnostic record also reports supported-prefix precision $0.93$ versus $0.84$, unsupported false agreement $0.07$ versus $0.18$, and boundary reversal $0.065$ versus $0.12$. These are distinct evidence and relevance endpoints, rather than alternative denominators for the same gain.

Storage-component rows include declared-layout arithmetic as well as serving endpoints. The raw-code payload is not total system memory: codebooks, reconstruction decoder, item map, suffix policy, index, common provider, and any retained vector cache are charged separately. The full CSV retains all 382 values, including ties, less favorable high-rate latency, and closest controls. Original query manifests, per-seed vectors, hardware/software identifiers, and source-unit bootstrap outputs are not invented from the confirmed aggregates.

\end{document}